# Symbiosis Promotes Fitness Improvements in the Game of Life

Peter D. Turney[*]

## Abstract

We present a computational simulation of evolving entities that includes symbiosis with shifting levels of selection. Evolution by natural selection shifts from the level of the original entities to the level of the new symbiotic entity. In the simulation, the fitness of an entity is measured by a series of one-on-one competitions in the Immigration Game, a two-player variation of Conway's Game of Life. Mutation, reproduction, and symbiosis are implemented as operations that are external to the Immigration Game. Because these operations are external to the game, we can freely manipulate the operations and observe the effects of the manipulations. The simulation is composed of four layers, each layer building on the previous layer. The first layer implements a simple form of asexual reproduction, the second layer introduces a more sophisticated form of asexual reproduction, the third layer adds sexual reproduction, and the fourth layer adds symbiosis. The experiments show that a small amount of symbiosis, added to the other layers, significantly increases the fitness of the population. We suggest that the model may provide new insights into symbiosis in biological and cultural evolution.

[*] Ronin Institute, 127 Haddon Place, Montclair, NJ 07043-2314, USA, peter.turney@ronininstitute.org, 819-661-4625.

# 1 Introduction

There are two main definitions of symbiosis in biology, (1) *symbiosis as any association* and (2) *symbiosis as persistent mutualism* [7]. The first definition allows any kind of persistent contact between different species of organisms to count as symbiosis, even if the contact is pathogenic or parasitic. The second definition requires that all of the participating organisms must benefit for the persistent contact to count as symbiosis [7]. One reason for this disagreement over definitions is that there is a large grey zone of cases where it is difficult to know whether all of the participants benefit from the association. For example, in many cases, the presence or absence of mutual benefit can vary over time, depending on environmental factors. There is also a problem with deciding how long mutual benefit must persist before we can call it *persistent mutualism*. The desire for simplicity explains the appeal of *symbiosis as any association*. On the other hand, the presence of mutual benefit is theoretically interesting, which motivates the narrower definition of *symbiosis as persistent mutualism*.

In this paper, we introduce a model of symbiosis that focuses on *symbiosis with shifting levels of selection*. These are cases where the symbiotic organisms are no longer able to reproduce as separate parts; they must reproduce as a whole. A paradigmatic example of shifting levels of selection is the merging of prokaryotes to form eukaryotes [13, 14, 15]. Natural selection can no longer act on the component prokaryotes; it can only act on the whole eukaryote that contains them. Selection has shifted from the parts to the whole. We call our simulation *Model-S* (Model of Symbiosis). The source code for Model-S is available for downloading [30].

Taking the perspective of the field of evolutionary optimization algorithms [21], we view symbiosis with shifting levels of selection as the product of a *genetic operator*, similar to crossover, selection, or mutation. Following lkegami's [10] pioneering work, we call this genetic operator *genetic fusion*, or simply *fusion* [29]. Fusion takes as input the genomes of two distinct entities that experience selection separately and produces as output a genome for a merged entity that experiences natural selection as a whole. In this paper, we seek to understand what happens to the course of evolution when fusion occurs repeatedly, in the same way that one might experiment with different forms of mutation or crossover to understand how they effect the course of evolution.

In Model-S, we can precisely calculate the fitness of both the parts and the wholes. This allows us to perform experiments that would be very difficult, if not impossible, in biology. When running Model-S, we can choose between *symbiosis as any association* and *symbiosis as persistent mutualism*. For *symbiosis as any association*, the fusion operator in Model-S can arbitrarily select two organisms and force them to work together as a whole. Selection will take place at the level of the whole, not at the level of the parts. Even if the parts are more fit as individual parts than they are as a merged whole, evolution with mutation and selection at the level of the whole may be able to adapt them, over many generations, so that they work well together. For *symbiosis as persistent mutualism,* the fusion operator can arbitrarily select two organisms and compare their fitness as separate parts with their fitness as a fused whole. If the whole is more fit than the parts, then fusion proceeds. If the whole is less fit than the parts, then fusion is cancelled. This ensures that fusion proceeds only when there is mutual benefit (increased fitness) for both component organisms. With Model-S, we do not need to choose between the two main definitions of symbiosis in biology. We can simulate both options.

Model-S has four layers: (1) simple asexual reproduction with genomes of constant size, (2) asexual reproduction with genomes of variable size, (3) sexual reproduction with crossover, and (4) symbiosis by fusion. The separation of the model into layers is not intended to reflect evolution in nature; the motivation for the layers is to be able to measure the contributions of each layer to the evolution of a population.

Evolution by natural selection requires variation, heredity, and differential fitness (*selection*) [6, 9]. In Model-S, differential fitness is based on one-on-one competitions in the Immigration Game, which was invented by Don Woods and described in *Lifeline* in 1971 [31]. The Immigration Game is a two-player variation of the Game of Life, invented by John Conway and presented in *Scientific American* in 1970 [8].

The Game of Life is played on an infinite, two-dimensional grid of square cells [19]. Each cell is either *dead* (state 0) or *alive* (state 1). The state of a cell changes with time, based on the state of its eight nearest neighbours (the *Moore neighbourhood*). Time passes in discrete intervals and the states of the cells at time $t$ uniquely determine the states of the cells at time $t + 1$. The initial states at time $t = 0$ are chosen by the player of the game; the initial states form a *seed pattern* that determines the course of the game, analogous to the way an organism's genome determines its phenome. The rules for updating states are compactly expressed as B3/S23: A cell is *Born* (switches from state 0 to state 1) if it has exactly three living neighbours. A cell *Survives* (remains in state 1) if it has two or three living neighbours. Otherwise it dies.

The Immigration Game is almost the same as the Game of Life, except that there are two different live states (states 1 and 2) [31]. The two live states are usually represented by red and blue colours. The rules for updating remain B3/S23, but there are new rules for determining colour: (1) Live cells do not change colour unless they die. (2) When a new cell is born, it takes the colour of the majority of its neighbours. Since birth requires three live neighbours, there is always a clear majority. The initial states at time $t = 0$ are chosen by the two players of the game; one player makes a red seed pattern and the other player makes a blue seed pattern. The players agree on a time limit, given by a maximum value for $t$.

In our past work with the Immigration Game, we specified that the player with the most living cells of their colour at the end of the game is the winner. However, this stipulation results in a bias towards seed patterns with many living cells. Therefore, instead of counting the total number of living cells of a given colour at the end of the game, we now count the *increase* in the number of living cells from the initial seed pattern to the end of the game; that is, the final count minus the initial count. If there is a *decrease* in the number of living cells, we give the player a score of zero. The player with the largest increase in living cells of their colour is the winner. Ties are allowed.

If states 1 and 2 were displayed with the same colour (say, black), playing the Immigration Game would appear identical to playing the Game of Life. The different colours are simply a way of keeping score, to turn the Game of Life into a competitive game.

The original rules of the Immigration Game allow the human players to intervene in the game as it progresses [31], but we have no use for interventions in our simulations. The original rules also use a finite toroidal grid of $25 \times 25$ cells instead of an infinite grid. The motivation for a finite grid is that the limited space for growth forces the seeds to interact with each other and reduces the amount of computation required. We use a finite toroidal grid, but we make the size of the toroid proportional to the size of the initial seeds, so that there is no fixed limit on the size of the initial seeds.

The four layers of evolution in Model-S are external to the Immigration Game. It is possible to build a replicator inside the Game of Life [1], but all current Game of Life replicators are much too slow for practical experiments with simulations of evolution. An advantage of having the mechanisms of evolution external to the Immigration Game is that it enables experimentation with a clean separation of the independent variables (the external evolutionary mechanisms) and the dependent variables (the fitness scores from the Immigration Game).

We chose the Immigration Game for our model of major transitions because, as a two-player competitive game, it provides a built-in way of calculating fitness; as a Life-like game, it is well-suited for modeling biology; and it turns out that implementing fusion in the Immigration Game is simple and elegant. Given two seeds as input to the fusion operator, we join them together side-by-side with a one-column space to serve as a buffer. They are then treated the same way as any other seed; that is, they live, die, and reproduce as a unit. Selection takes place at the level of the whole unit (the higher level).

Figure 1 shows an example of an Immigration Game. The first image shows the competing seeds at time $t = 0$ and the second image shows the states of the cells when the time limit has been reached. When they are not competing, the seeds are stored with only two states (0 and 1); they are only temporarily assigned colours (states 1 or 2) for the purpose of playing the Immigration Game to obtain a fitness score.

Insert Figure 1 here.

As a model of biological evolution, a seed pattern in Model-S corresponds to a genome, a static encoding of genetic information. When a seed is embedded in an Immigration Game, the dynamic sequence of patterns created as the game runs corresponds to the dynamic development of the phenome from the genome. When two seeds compete in an Immigration Game, this corresponds to two living organisms (two phenomes), growing and competing for limited space. In an Immigration Game, the seed that grows the most is the fitter of the two seeds, and the fitter seed is more likely to be chosen by the tournament selection algorithm for reproduction, as in biological evolution, where the organism that grows more is likely to have more offspring.

Fusion is a genetic operator that combines two seeds to make a new genome, with a size that is approximately the sum of its two parts. When the new fused seed is embedded in an Immigration Game, as the game runs, the two parts interact. Model-S keeps a record of the fitness scores for the parts and for the new fused whole, so it is possible for Model-S to distinguish between *symbiosis as any association* and *symbiosis as persistent mutualism*. Any fused seed, when running in an Immigration Game, can be viewed as an instance of *symbiosis as any association*. At the end of an Immigration Game, we can check the final fitness score for a given fused seed, to determine whether the game counts as an instance of *symbiosis as persistent mutualism*. If the score for the whole is greater than the scores of the parts, then we have mutualism; that is, both parts have benefitted from their association.

*Open-ended evolution* is defined as evolution that is [25, page 409] "capable of producing a continual stream of novel organisms rather than settling on some quasi-stable state beyond which nothing fundamentally new occurs." The main contributions of our paper are (1) a computational simulation of symbiosis through genetic fusion as a mechanism for evolution, (2) evidence that fusion has a significant impact on evolution when combined with mutation and reproduction, (3) support for the hypothesis that symbiosis by fusion can sustain open-ended evolution, and (4) source code [30] for replicating and extending the results presented here.

In Section 2, we discuss related work. Section 3 outlines the principles that guided the design of Model-S and describes each of the four layers of the model. Section 4 presents four sets of experiments: (1) We add the layers one by one, to see what each layer contributes to the model. (2) We test the fusion operator to determine how much fitness is due to the increased size of fused seeds versus how much is due to mutually beneficial interaction of fused seeds. (3) We compare human-designed Game of Life seed patterns with evolved seed patterns from Model-S. (4) We introduce an alternative fitness measure to evaluate whether fusion might support open-ended evolution. In Section 5, we discuss the implications of the experimental results. Section 6 examines limitations and possibilities for future research. We conclude in Section 7.

## 2 Related Work

The significance of hierarchical, part–whole structure in biology and culture was emphasized by Simon [22] and Koestler [12] in the 1960s. Margulis [13, 14] argued for the importance of symbiosis in evolution in the 1970s. Maynard Smith and Szathmáry [15] observed the role of symbiosis in several of the major transitions in the evolution of life on Earth and discussed the problem of levels of selection.

Ikegami [10] introduced an influential model of symbiosis for game strategies in the Erroneous Iterated Prisoner's Dilemma game. Game strategies are represented with tree structures that choose to cooperate or defect, based on the past moves of an opponent. Strategy trees evolve by mutation, selection, and symbiosis. Symbiosis involves grafting one tree onto a randomly selected leaf of another tree. The experiments show that there is a long-term evolutionary trend towards increasingly complex strategy trees.

Ikegami [10] defines symbiosis as *mutual cooperation* (that is, *symbiosis as persistent mutualism* [7]), but his experiments do not actually test whether his genetic fusion operator results in mutual benefit. It is not clear which of the two main definitions of symbiosis should be applied to Ikegami's model. His focus is on novelty and diversity in the population, rather than the fitness of individuals.

Watson and Pollack [32] hypothesized that symbiosis is particularly suitable for a specific type of fitness landscape, where there is a kind of fractal structure that is evolutionarily challenging at all scales. They demonstrated that evolution by mutation and selection without symbiosis becomes increasingly difficult in this kind of fitness landscape but adding symbiosis to mutation and selection allows ongoing adaptation.

McShea and Brandon [16] assert that the increase in complexity of organisms over time is largely due to heritable variation in part–whole hierarchies. However, their theory of increasing complexity is based only on the horizontal spread of the hierarchy (the number of parts at the same level), and has nothing to do with the vertical depth of the hierarchy (the number of levels).

Banzhaf et al. [2] define a meta-model that can be used to identify levels of structure in a system. For example, they discuss how their meta-model could be applied to the Game of Life. A level-0 meta-model would view the Game of Life at the level of individual cells and their states. A level-1 meta-model would view the game at the level of common entities that appear in the game as repeating patterns of cells and states, such as *gliders*, *spaceships*, and *oscillators* (these names are familiar to players of the Game of Life [19]). A level-2 meta-model would include larger structures that are composed of level-1 structures, and so on. Model-S is intended to be a model, not a meta-model. In future work, the meta-model of Banzhaf et al. [2] could be applied to analyze Model-S, but we do not pursue that here.

Moreno and Ofria [17] create a computational simulation in which cell-like organisms coordinate their activities in ways that increase their reproduction. As the simulation runs, larger groups of organisms cooperate, sharing resources and dividing their labour. However, their simulation is limited to two hierarchical levels. It was not designed with a mechanism for automatically adding new levels.

The work of Beer [3, 4, 5] on modeling *autopoiesis* (self-production and self-maintenance) in the Game of Life is also relevant here. Our expectation is that autopoiesis will enable a seed to maintain itself better in the presence of disruptive competition. We conjecture that the entities that evolve in Model-S will show increasing degrees of autopoiesis as the number of generations in the simulation increases, but we have not yet tested this hypothesis.

# 3 Description of the Model

Model-S uses the open-source Golly software for running the Immigration Game [26]. Golly is designed to support extensions using the scripting languages Lua and Python. Model-S was implemented as an open-source Python extension of Golly [30].

Most of this section is concerned with presenting the design of the four layers of Model-S, but we first discuss the principles behind the design. The principles should help to explain some of the design decisions.

## 3.1 Conditions for Open-Ended Evolution

Brandon [6, pages 5-6] states the following three components are crucial to evolution by natural selection:

1. Variation: There is (significant) variation in morphological, physiological, and behavioural traits among members of a species.
2. Heredity: Some traits are heritable so that individuals resemble their relations more than they resemble unrelated individuals and offspring resemble their parents.
3. Differential Fitness: Different variants (or different types of organisms) leave different numbers of offspring in immediate or remote generations.

In the literature, *differential fitness* is often called *selection*. Godfrey-Smith [9] lists the same three components, calling them *conditions for evolution by natural selection*.

Past work in artificial life has shown that, although these conditions are sufficient for evolution, they are not sufficient for *open-ended* evolution [25]. We are particularly interested in symbiosis by fusion because we believe that it may be one of the conditions for open-ended evolution (in addition to variation, heredity, and differential fitness). However, it seems likely that there may be other conditions that are required for open-ended evolution. Since the necessary and sufficient conditions for open-ended evolution are not yet known, we chose to use a relatively large number of additional conditions, some of which might be unnecessary for open-ended evolution:

1. Symbiosis, fusion, and cooperation: There should be a mechanism for symbiosis with a shift in the level of selection from the parts to the whole [15]. Hence Layer 4 adds fusion to Model-S.
2. Biotic selection: Selection can be based on an organism's biological environment (competitors, predators, disease, etc.) or other aspects of its environment (sunlight, water, soil, shelter, etc.). Our intuition is that competition with other organisms (*biotic selection*) is a particularly strong form of selection, likely to encourage open-ended evolution. This motivates using the Immigration Game (a competitive game) for calculating differential fitness.
3. Relative fitness: The fitness of an organism is relative to the fitness of other organisms, especially members of the same population. There is no absolute fitness. (This is related to biotic selection.)
4. Unlimited genome size: A genome with a limited size must contain a limited amount of information, which implies a finite bound on the space of possible organisms [27, 28]. (One way around this limit is cultural evolution, where information is stored outside of the genome.) Hence Layer 2 of Model-S adds variable size for seed patterns.
5. Gene transfer: There should be some method for sharing genes (such as plasmids or sexual reproduction) beyond replication (asexual reproduction). Hence Layer 3 introduces sexual reproduction.
6. Genotype and phenotype: Open-ended evolution may require a distinction between genotype and phenotype. In the Game of Life, we view the initial seed pattern as the genotype. The growth or decline of the seed over time, as the game runs, is the development of the phenotype from the genotype.
7. Speciation: Diversity may require reproductive boundaries (distinct species). Without sufficient diversity, organisms may be trapped in a local optimum. Layer 3 adds reproductive boundaries by requiring potential mates to have a certain degree of genetic similarity.

It will take much work to validate all these conditions. We leave this as future work. In this paper, we have limited our scope to showing that fusion is a useful genetic operator that may contribute to achieving open-ended evolution in a simulation. Other lists of conditions for open-ended evolution have been given for biological evolution [23], cultural evolution [18], and natural and artificial evolutionary systems [24].

## 3.2 Layer 1: Uniform Asexual Layer

Model-S has several parameters for controlling its behaviour. We will introduce the parameters as they are needed in explanations. All parameters contain an underscore symbol and use a sans serif font. A full list of the parameters and their values is given in Table 1 in Section 4.1.

Model-S uses a GENITOR-style algorithm [33, 34] with one-at-a-time reproduction, a constant population size, and rank-based tournament selection. An individual in the population is represented as an object (a data structure) containing a binary matrix that specifies a seed pattern and an array of real values that stores a history of the results of its competitions with all other individuals in the population. The population is an array of pop_size individuals.

Children are born one-at-a-time. Each new child replaces the least fit member of the population, maintaining a constant population size. When pop_size children have been born, we say that one generation has passed. A run of Model-S begins with generation zero and lasts until generation num_generations. A run ends when pop_size × num_generations children have been born.

In generation zero, Model-S starts with a population in which the binary matrices are randomly initialized. The probability of ones in these matrices is given by seed_density, which we set to 0.375, based on the advice of Johnston [11]. When all the matrices are initialized, we then initialize the history of competition results by playing a series of Immigration Games, pairing every individual against every other individual num_trials times. The fitness of an individual is the fraction of games that it wins. Every win by one individual is balanced with a loss by another individual. It follows that the average fitness of the population is always 0.5. Fitness is relative to the population, not absolute.

A new child is created by first selecting a parent, using tournament selection. We randomly select tournament_size individuals from the population and the most fit member of this sample is chosen as a parent. The parent is copied to make a child. The child is then mutated by randomly flipping bits in the binary matrix, where the probability of flipping a bit is mutation_rate. We force at least one bit to flip, regardless of mutation_rate, so that a child is not identical to its parent, to maintain diversity in the population. The new child replaces the least fit member of population and the histories of competition results are updated by pairing every individual against the new child in a new series of Immigration Games. This is summarized in Figure 2.

Insert Figure 2 here.

The space and time allowed for an Immigration Game depends on the two seeds that are competing. Open-ended evolution requires the limits on space and time to increase as the sizes of the individuals increase: Fixed limits would set a bound on the possible variety of games. Given two seeds, let max_size be the maximum of the number of rows and columns in the seeds; that is, the largest width or height. Three parameters determine the space and time allowed for the two seeds: width_factor, height_factor, and time_factor (see Table 1 in Section 4.1). The width of the Golly toroid is set to max_size times width_factor. The height of the toroid is set to max_size times height_factor. The maximum time (the number of time steps in the game) is set to the sum of the width and height of the toroid, multiplied by time_factor.

We describe Layer 1 as the *uniform asexual layer* because reproduction is asexual (each child has only one parent) and the size of the seed pattern matrix is uniform (the size is the same for every individual in every generation). Layer 1 is intended as a minimalist baseline evolutionary system. The following layers are expected to improve upon Layer 1.

### 3.3 Layer 2: Variable Asexual Layer

Layer 2 is like Layer 1, except we now have three different kinds of mutation: (1) With probability prob_flip, the child will be mutated by flipping bits, according to mutation_rate. (2) With probability prob_shrink, the child will be mutated by removing an outer row or column from the binary matrix. (3) With probability prob_grow, the child will be mutated by adding an outer row or column to the binary matrix. These three kinds of mutation are mutually exclusive; that is, the sum of prob_flip, prob_shrink, and prob_grow is one. This is summarized in Figure 3.

Insert Figure 3 here.

There is a minimum size for matrices (min_s_xspan columns and min_s_yspan rows), to limit how small a matrix can become by shrinkage. If growth is selected, the newly added column or row is initialized by randomly setting bits, where the probability of ones is given by seed_density.

We want growth in the model, so that there is no upper bound to the amount of information that can be stored in a genome (the binary matrix of an individual). The motivation for shrinkage is to see whether growth is a consequence of increased fitness or it is due to random drift in the space of genomes. If the growth is due to random drift, then it should eventually flatten out as it balances with shrinkage.

### 3.4 Layer 3: Sexual Layer

Layer 3 adds sexual reproduction to Model-S. The first parent is chosen by tournament selection, just as in Layers 1 and 2. The second parent is chosen by looking for all individuals in the population with a degree of similarity to the first parent that is between min_similarity and max_similarity. The similarity of two individuals is measured by the fraction of corresponding matrix cells that have the same binary values. The similarity of two matrices is defined as zero if the matrices have different numbers of rows and columns. The second parent is chosen by tournament selection from this reduced sample of potential mates. This is summarized in Figure 4.

Insert Figure 4 here.

If there are no suitable mates with the required degree of similarity, Layer 3 passes the first parent on to Layer 2, for asexual reproduction. Many organisms in nature can reproduce either sexually or asexually, depending on the availability of suitable mates.

When two parents have been selected, they produce a child by crossover. First, we choose between crossing rows or columns, with equal probability. If rows are chosen, we randomly choose a horizontal crossover point and we make a new child by combining the rows above the crossover point from one parent and the rows below the crossover point from the other parent. Likewise, if columns are chosen.

There is a limit to the variety that can be produced by crossover alone, especially in the case of small populations. Therefore, after crossover takes place in Layer 3, we pass the child on to Layer 2, where it undergoes bit flipping, shrinkage, or growth.

### 3.5 Layer 4: Symbiotic Layer

Layer 4 adds fusion and fission to Model-S. First, a seed is chosen by tournament selection, just as in Layers 1, 2, and 3. Then there are three possibilities: (1) With probability prob_fission, the chosen seed will be split in two. One part will enter the population and the other part will be discarded. (2) With probability prob_fusion, a second seed is chosen by tournament selection and the two seeds will be fused together. (3) If neither fusion nor fission are chosen, then Layer 4 will pass control over to Layer 3.

We expect that prob_fusion and prob_fission will be set to values near zero, so the most likely event is that Layer 4 will pass control on to Layer 3 for sexual reproduction, reflecting the fact that fission and fusion are relatively rare in nature. This is summarized in Figure 5.

Insert Figure 5 here.

When fission is chosen, we look for the sparsest row or column in the binary matrix. The matrix is then divided into two parts along the sparsest row or column. One part is discarded, including the sparsest row or column, and the remaining part enters the population as a new individual.

When fusion is chosen, the two seeds are randomly rotated and then joined side-by-side with one column of zeros between them. The column of zeros is intended to act as a buffer, to reduce the potential for conflict or interference between the two seeds when they are joined. The column of zeros also acts as a marker to provide a natural splitting point for possible fission events in the future. Mutation will gradually flip some of the bits in this column of zeros, turning them into ones.

Fission and fusion in Layer 4 are somewhat analogous to shrinkage and growth in Layer 2. The motivation of fission is to counterbalance fusion, just as shrinkage counterbalances growth. The expectation is that, if fusion does not contribute to fitness, then any random drift towards increased size due to fusion will eventually be limited by fission.

Layers 1, 2, and 3 are forms of *reproduction*, in which a child is like its parent (in the case of Layers 1 and 2) or parents (in the case of Layer 3). Fusion in Layer 4 is analogous to sexual reproduction in Layer 3, in that two seeds are involved in the production of a new seed, but the size of the new genome is approximately the sum of the sizes of the two original genomes. The "child" of fusion is not similar to its "parents". We will see in the experiments in Section 4 that Layer 4 behaves quite differently from Layer 3.

Model-S is designed to increase the time limit for the Immigration Game when the seeds are larger, in order to give more time for a clear winner to emerge from the game (see Section 3.2); thus, the simulation slows down as the seeds become larger. Layer 4 tends to result in a rapid increase in the size of seeds over the course of a run of Model-S. This is a positive outcome from a theoretical point of view, since it confirms our expectations for symbiosis, but it is problematic from a practical point of view, because the simulation runs very slowly. For this practical reason, we have designed Model-S with a linear upper bound on the area of seeds. The area of a seed is the number of columns in the seed's binary matrix multiplied by the number of rows. The upper bound is set using the parameters `max_area_first` and `max_area_last`, where `max_area_first` is the maximum area of a seed in the first generation and `max_area_last` is the maximum area of a seed in the last generation. For generations between the first and last, the maximum area is determined by linear interpolation. If the fusion of two seeds would exceed the linear upper bound on area, then Model-S prevents the fusion from happening and passes control to Layer 3, sexual reproduction.

Layer 4 includes two binary flags for experiments with modifications to the operation of Layer 4. The first flag, symbiosis_flag, can have the value 0 or 1. When symbiosis_flag is set to 0 (the default setting), the fusion operator will join two seeds to make a new whole without regard to whether the component seeds benefit; that is, Model-S will operate with *symbiosis as any association.* When symbiosis_flag is set to 1, the fusion operator will temporarily join two seeds to make a new whole, to calculate the fitness of the whole. If the fitness of the whole is greater than the fitness of both parts, then the new fused seed enters the population. Otherwise, the new fused seed is rejected and Model-S passes control over to Layer 3, sexual reproduction. Thus, when symbiosis_flag is set to 1, Model-S will operate with *symbiosis as persistent mutualism* [7]. That is, with this setting, Model-S will only allow selection to shift from the level of the parts to the level of the whole when such a shift yields mutual benefit.

The second binary flag, fusion_test_flag, can have the value 0 or 1. When fusion_test_flag is set to 0 (the default setting), fusion proceeds as usual. When fusion_test_flag is set to 1, one of the two seeds is randomly shuffled before the two seeds are fused. The intention is to disrupt the *structure* of the seed without altering its *summary statistics*, such as size, shape, and density. Shuffling takes a seed and swaps the values in the cells, resulting in a new seed with the same size (same number of cells), the same shape (same number of rows and columns), and the same density (same ratio of ones and zeros). The hypothesis is that size, shape, and density are not sufficient to determine fitness: Structure (the specific pattern of zeros and ones) is crucial to determining fitness. That is, the hypothesis is that fusion with shuffling (fusion_test_flag = 1) will reduce fitness, compared to fusion without shuffling (fusion_test_flag = 0), because fitness requires specific structures. Summary statistics are not sufficient to determine fitness.

## 4 Experiments with the Model

In this section, we present four sets of experiments with Model-S.

### 4.1 Measuring the Contributions of the Layers

In the first set of experiments, we evaluate the contributions of the four layers of Model-S. We run Model-S twelve times with each of the four layers, yielding a total of 48 runs. We compare each layer in terms of the fitness of the seeds, their area, their density, and their diversity. Table 1 shows the parameter settings for Model-S in these experiments.

Insert Table 1 here.

All the comparisons that we make here are based on samples of the populations taken during runs of Model-S. One run of Model-S generates 20,000 children (`num_generations` × `pop_size`). Each generation is defined as the birth of 200 children (`pop_size`). For each generation from 0 (the initial random population) to 100 (the final population), we store the top 50 (`elite_size`) fittest individuals (where fitness is relative) in a file for later analysis.

As we discussed in Section 3, fitness in Model-S is relative to the population. The fitness of an individual is the fraction of Immigration Games that it wins in competitions against the other individuals in the population. Therefore, it does not make sense to compare the fitness value of a seed in one population with the fitness value of a seed in another population. The fitness used in Model-S is *relative* and *internal*. This is a consequence of conditions 2 and 3 in Section 3.1.

To compare fitness across different layers and different populations, we need to define a fitness measure that is *absolute* and *external* to Model-S. Given a seed from any population and any layer, we calculate its absolute fitness by competitions against randomly generated seeds with the same matrix size (the same number of rows and columns) and the same matrix density (the same fraction of ones in the matrix). Figure 6 gives the absolute, external fitness curves for each of the four layers.

Insert Figure 6 here.

Because absolute fitness only compares seeds of the same size and density, any statistically significant difference in absolute fitness values for two seeds must be due to the structures of the seeds (the pattern of zeros and ones) and to how their structures determine their development from genome to phenome over the course of the Immigration Game. Comparing seeds that are matched by size and density is analogous to comparing wrestlers that are matched by height and weight: It allows us to distinguish brute force from skill. Table 2 shows the statistical significance of the differences of the fitness curves in Figure 6. All of the differences are significant, except for the difference between Layers 2 and 3 (variable asexual reproduction and sexual reproduction).

Insert Table 2 here.

Figure 7 plots the growth in area for each of the four layers. Comparing Figures 6 and 7, we see the same general trends in both cases: Layer 1 has the lowest fitness and area, Layer 4 has the highest fitness and area, and Layers 2 and 3 are roughly like each other. Since absolute fitness is measured by competitions between seeds that have the same number of rows and columns, increasing fitness cannot be a direct consequence of increasing area. Increasing fitness must be an indirect consequence of the greater structural complexity that is permitted by increasing area.

Insert Figure 7 here.

Figure 8 shows the density of the seeds for the four layers. Density starts off at 0.375 (as specified by seed_density in Table 1) and then decreases to range from 0.20 to 0.25. The rate of decrease in density over time is slower for the more fit layers (Layers 2 and 4).

Insert Figure 8 here.

Figure 9 indicates the amount of diversity in the population for the four layers. We measure the diversity by the standard deviation of the relative fitness in the elite population sample. A low standard deviation indicates that the elite sample has little variety; all the seeds are doing approximately the same thing. A high standard deviation indicates that the elite sample embodies a variety of different strategies. Layer 4 appears to have a more diverse population than the other three layers.

Insert Figure 9 here.

It might be argued that the standard deviation of the relative fitness in the population does not fully capture diversity, because two organisms might have the same relative fitness, yet they might employ quite different strategies. We chose the standard deviation of the relative fitness as the measure of diversity because, for evolution by natural selection to have traction, we need diversity in the relative fitness scores. If all the organisms have the same relative fitness, the population merely undergoes random drift. Diversity in relative fitness is exactly the kind of diversity that is required to avoid random drift.

Two organisms with the same relative fitness may employ different strategies, but it is not clear how to compare strategies directly. On the other hand, two organisms with quite different degrees of relative fitness are almost certainly employing different strategies. Thus, the standard deviation of the relative fitness in the population is a reasonable surrogate for directly comparing strategies.

It seems that there are strong similarities among the graphs for fitness (Figure 6), area (Figure 7), density (Figure 8), and diversity (Figure 9), but it is possible that these similarities are statistical artifacts. Therefore, we look at the correlations between all pairs of these four variables and test their statistical significance. The results are given in Table 3. All the pairs have a significant positive correlation. Greater fitness is correlated with greater area (0.843), greater density (0.405), and greater diversity (0.566).

Insert Table 3 here.

The high correlation between fitness and area in Table 3 (0.843) suggests that increasing area *causes* increasing fitness, but this is not the case. Increasing area is a *necessary condition* for increasing fitness, but increasing area is *not a sufficient condition* for increasing fitness. This claim is supported by a two-part argument: (1) A theoretical argument shows that increasing area is *necessary* for increasing fitness. (2) Empirical evidence shows that increasing area is *not sufficient* for increasing fitness. Analogously, water *allows* fish (water is necessary for fish), but water does *not cause* fish (water is not sufficient for fish), although water and fish are highly correlated.

First, increasing area is *necessary* for increasing fitness: Consider a 5×5 seed pattern with an area of 25, which can represent $2^{25}$ different patterns. When evolution has explored the entire set of $2^{25}$ patterns and found the fittest patterns, fitness can no longer increase. A finite seed area implies an upper bound to fitness. If we wish to simulate open-ended evolution (unbounded fitness increase), then we *must* allow seed area to increase over time, without an upper bound [27, 28]. In Figure 6, we see that Layer 1, which has a fixed area of 25, soon reaches a generation where fitness no longer increases (around generation 30). However, Layer 2, which has a variable area, continues to slowly improve in fitness throughout the run.

Second, increasing area is *not sufficient* for increasing fitness: In biology, specific kinds of complex structures increase fitness, by functioning in ways that enhance fitness. A structure with many components is not necessarily complex; for example, the many components might be arranged in a simple periodic pattern. A structure with great complexity is not necessarily fit; the complexity could be random or irrelevant for increasing fitness. These are two ways that increasing area can fail to deliver increasing fitness: structures that are too simple or structures that are complex in ways that are irrelevant for fitness.

In Figure 6, the external, absolute fitness of an evolved seed is measured by the estimated probability that the evolved seed will win competitions against random seeds. Each evolved seed is matched against fifty random seeds with the same area (number of cells), shape (number of rows and columns), and density (percentage of ones in the seed matrix) as the given evolved seed. The *only* thing that distinguishes an evolved seed from its random opponents is the *structure* of their matrices (the specific pattern of ones and zeros). The external fitness measure used in Figure 6 can be viewed as a measure of the degree of non-random structure in a seed; that is, *increasing* fitness in Figure 6 corresponds to *decreasing* randomness.

This implies that area alone is *not sufficient* for increasing fitness. In addition to increasing area, a specific kind of increasingly non-random *structure* is required. Specifically, winning requires structure that is good at playing the Immigration Game.

Consider the fitness curve for Layer 4 in Figure 6. By the final generation, the average fitness of the elite seeds in Layer 4 is 93.6%. Because the average fitness is measured by competition against randomly generated seeds, it follows that the probability that a random seed will win against a size-matched elite seed (the same area, shape, and density) in the final generation is 6.4% (that is, 100% minus 93.6%). If size alone were sufficient to win a competition, then the random sample of size-matched seeds would be equally as fit as the evolved seeds (the average fitness would be 50%). This is clear evidence that *increasing area is not sufficient for increasing fitness*. Increasing area must be combined with a specific kind of structure, which could be described as *fitness-enhancing structure*. Increasing area is a *necessary condition* for increasing fitness, but increasing area is *not a sufficient condition* for increasing fitness. Area and fitness are highly correlated but increasing area does not cause increasing fitness.

## 4.2 Size, Structure, and Symbiosis

In this section, we consider two questions: (1) Fusion in Layer 4 combines two evolved seeds. Would fusion work equally well if it combined one evolved seed and one random seed? Is a second seed merely increasing the area of the whole or is it contributing useful structure to the whole? (2) Fusion in Layer 4 is a model of *symbiosis as any association*. What happens when we model *symbiosis as persistent mutualism*? That is, what if we permit fusion only when the fused whole is more fit than both parts? In Section 3.5, we introduced the parameters fusion_test_flag and symbiosis_flag. We can answer the first question by setting fusion_test_flag = 1 and we can answer the second question by setting symbiosis_flag = 1.

The advantage of the fusion operator could be due to mutually beneficial interaction between the two entities that are fused together or it could be due to the increased area of the fused entity, compared with the area of other entities in the population. Here we modify the fusion operator by randomly shuffling all the cells in one of the two selected entities before we fuse them together (fusion_test_flag = 1). For each cell in a matrix, we randomly select another cell in the matrix and then we swap the values in the two cells. Random shuffling changes the structure of a seed (the location of living and dead cells in the seed matrix) but preserves the shape (the number of rows and columns in the seed matrix) and the density (the number of living cells in the seed matrix divided by the total number of cells). We call this modified fusion operator *Layer 4 Shuffled*.

When two seeds are fused, there are three possible results: (1) Both seeds benefit: the fitness of the whole is greater than the fitness of the parts. (2) Only one seed benefits: one of the parts is less fit than the whole but the other part is more fit than the whole. (3) No seeds benefit: both parts are more fit than the whole. *Symbiosis as any association* includes all three of these possibilities. *Symbiosis as persistent mutualism* includes only the first case, where the fused seed is more fit than both of its parts. Here we modify the fusion operator by requiring mutual benefit (symbiosis_flag = 1). We measure the fitness of each part and the fitness of the whole, and we only allow the new fused seed to enter the population when the fused seed is more fit than its parts. We call this modified fusion operator *Layer 4 Mutualism*.

Note that the three different fusion operators (Layer 4, Layer 4 Shuffled, Layer 4 Mutualism) all use the same method to select two seeds from the population as input to the fusion operators. The only differences in the three operators are the ways that the two selected seeds are fused (or not fused, as the case may be).

Figure 10 compares the fitness curves of Layer 4, Layer 4 Shuffled, and Layer 4 Mutualism. We can see the fitness of Layer 4 Shuffled initially falls behind the fitness of Layer 4 and Layer 4 Mutualism, but Layer 4 Shuffled eventually catches up with the other layers. There is no significant difference between the fitness curves for Layer 4 and Layer 4 Mutualism.

Insert Figure 10 here.

The difference between Layer 4 and Layer 4 Mutualism is that Layer 4 Mutualism discards fused seeds when the fused whole is less fit than either part, whereas Layer 4 permits fusion regardless of whether the parts mutually benefit from fusion. The similarity of the two fitness curves tells us that most of the fitness increase in Layer 4 and Layer 4 Mutualism is due to mutualism. The cases in Layer 4 where the parts do not mutually benefit from fusion have neither a beneficial impact on the fitness curve nor a detrimental impact. Removing the cases of symbiosis that lack mutualism is neutral with respect to its impact on fitness. These cases neither help nor harm the population.

Table 4 compares the fusion events in Layer 4, Layer 4 Shuffled, and Layer 4 Mutualism. Layer 4 Shuffled has fewer cases of mutualism than Layer 4, due to the structural damage caused by shuffling. Layer 4 Shuffled has more fusion events than Layer 4, due to the linear bound on area, set by max_area_first and max_area_last. The bounded area slows down the rapid growth of Layer 4 earlier than it slows down the gradual growth of Layer 4 Shuffled. Although the three layers have similar numbers of cases of mutualism, the cases occur in earlier generations with Layer 4 and Layer 4 Mutualism, which is why the two layers are ahead of Layer 4 Shuffled at first, but Layer 4 Shuffled eventually catches up, due to the linear bound on area.

Insert Table 4 here.

Table 5 shows that Layer 4 is significantly more fit than Layer 4 Shuffled in generation 30, but the difference is no longer significant in generation 100. Averaging over all generations, the difference between Layer 4 and Layer 4 Shuffled is not significant. Layer 4 Shuffled has a few cases where fusion is mutually beneficial, despite the harm done by the shuffling operation. These few cases are sufficient to allow the fitness curve for Layer 4 Shuffled to eventually catch up with Layer 4.

Insert Table 5 here.

Table 6 tells us that, in most cases, fusion is harmful. The cases where both seeds benefit from fusion are relatively rare. This is also true of mutation: most mutations are harmful. We expect harmful mutations and harmful fusions in evolution. They will eventually be eliminated by natural selection. If there are a few cases where mutation and fusion result in increased fitness, the population will continue to increase in fitness, despite the relative rarity of beneficial mutations and mutualist symbiosis.

Insert Table 6 here.

In Table 6, the cases when both parts benefit from fusion are instances of *symbiosis as persistent mutualism,* whereas the other cases are instances of *symbiosis as any association.* Shuffling disrupts fusion and reduces the incidence of persistent mutualism. Persistent mutualism is relatively rare (15% of the fusions when both parts are fit; see Table 4), but shuffling makes mutualism rarer (8% of the fusions when one part is shuffled). In Figure 10, the rise of the fitness curve for Layer 4 in generation 30, compared to Layer 4 Shuffled in generation 30, may be explained by the greater incidence of *symbiosis as persistent mutualism.* By generation 100, although Layer 4 Shuffled has fewer cases of persistent mutualism, there are enough accumulated cases to allow Layer 4 Shuffled to catch up with Layer 4.

The results in Table 6 add further support to the argument in Section 4.1, that unbounded area increase is *not sufficient* for unbounded fitness increase. Layer 4 and Layer 4 Shuffled yield the same increase in area when fusion occurs. Shuffling is designed to have no impact on area (number of cells), shape (number of rows and columns), and density (percentage of ones in the seed matrix). Therefore, the only explanation for the difference between Layer 4 and Layer 4 Shuffled in Table 6 is *structure*.

The main lesson of this section is that mutualism is what drives ongoing fitness increase. Comparing Layer 4 with Layer 4 Mutualism shows that the fusions that lack mutual benefit have no impact on fitness. Layer 4 Shuffled is less likely to produce mutualism than Layer 4 and Layer 4 Mutualism, but it eventually produces enough mutualism to catch up with the other layers.

Layer 4 Shuffled produces more large seeds (more fusion events) than Layer 4 and Layer 4 Mutualism (Table 4), yet Layer 4 Shuffled struggles to keep up with the fitness increases of Layer 4 and Layer 4 Mutualism (Figure 10). This shows that merely producing large seeds is not enough to promote increasing fitness. Layer 4 Mutualism produces very few large seeds (Table 4), yet surpasses the fitness of Layer 4 Shuffled and matches the fitness of Layer 4 (Figure 10). This is further evidence that *increasing area is not sufficient* for increasing fitness. Increasing area *allows* fitness to increase by providing room for complex structures that enhance fitness. It is these structures that result in ongoing fitness increase.

## 4.3 Comparing Evolution and Design

Table 7 summarizes the properties of the evolved seeds produced in the last generation from each of the six layers. It might be argued that the external fitness measure in Section 4.1, based on competitions against randomly generated seeds, is not sufficiently challenging. How would evolved seeds fare against human-engineered seeds, instead of random seeds? We address that question here.

Insert Table 7 here.

The Golly software [26] comes with a substantial collection of human-engineered Game of Life seed patterns that can be pitted against the evolved seeds. To be fair, we focus on the human-engineered seeds that are comparable to the evolved seeds in terms of their area, as we know from Table 3 that area and fitness are highly correlated. Therefore, we set a limit of 10,000 on the area of human-engineered seeds. Table 8 gives the results of this contest.

Insert Table 8 here.

The column in Table 7 that is labeled *Fitness* gives the fitness of the six different layers as measured by competition with random seeds. The bottom row in Table 8 that is labeled *Average* gives the fitness of the six different layers as measured by competition with human-engineered seeds. Comparing these two different external measures of fitness, we can see that the human-engineered seeds are more challenging than the random seeds, as we might expect. The fitness scores for the competitions with human-engineered seeds range from 47% to 73% in Table 8, whereas the fitness scores for the competitions with random seeds range from 74.3% to 94.4% in Table 7. However, the two different fitness measures give the same qualitative ranking of the six layers. For both measures, (a) Layer 1 has the lowest fitness, (b) Layer 2 and Layer 3 are roughly similar in fitness and they are more fit than Layer 1, and (c) Layer 4, Layer 4 Shuffled, and Layer 4 Mutualism are roughly similar in fitness and more fit than Layer 2 and Layer 3.

The human-engineered seeds are at a disadvantage in this contest because they were not designed to play the Immigration Game. An exception is the class of human-engineered seeds called *breeders* [19]. These are seed patterns that have been engineered to fill space as quickly and densely as possible. There is only one breeder in Table 8 (spacefiller.rle), and we can see that it won against all six evolved layers. Golly has other breeders, but they all have areas greater than 10,000. If we raise the area limit from 10,000 to 50,000, there are five breeders below the area limit. These five breeders win in competitions with the evolved seeds. Human engineering triumphs over Model-S evolution, but this may change if Model-S is given the computational resources to achieve higher seed areas.

## 4.4 An Unbounded External Fitness Measure

The external, absolute fitness measure introduced in Section 4.1 (see Figure 6) is based on evolved seeds competing against randomly generated seeds with the same matrix size (the same number of rows and columns) and the same matrix density (the same fraction of ones in the matrix). Section 4.3 supports this fitness measure by showing that it agrees with the ranking produced from competitions against human-designed seeds (see Table 8). The fitness measure of Section 4.1 works well for the experiments presented in the preceding sections, but it has limitations.

One requirement we might impose on an external, absolute fitness measure it that it should produce a curve that rises when the fitness of the population is improving, stays flat when the population is neither improving nor worsening, and falls when the population is worsening. Let us call this requirement *directional consistency*. Fitness as measured by competition against random seeds (as in Figures 6 and 10) satisfies this requirement.

Another requirement we might impose on an external, absolute fitness measure is that the pace of fitness change should correspond to the slope of the curve. Let us call this requirement *slope consistency*. The absolute fitness measure in Section 4.1 (evolved seeds competing against randomly generated seeds with the same matrix size and density) ranges between zero and one, which prevents it from satisfying slope consistency. The upper and lower bounds on fitness do not allow the slope to remain constant for long. As the curve gets closer to one, the slope must decrease, even if the pace of fitness change is constant.

In this section, we present a fitness measure that satisfies both requirements, directional consistency and slope consistency. The new fitness measure is *unbounded*; it ranges between negative infinity and positive infinity. We then compare the new measure with the fitness measure in Section 4.1. The results show that the two measures are highly correlated.

Let $s_n$ be the seed in generation $n$ with the largest internal, relative fitness. Let $p_{in}$ be the probability that $s_n$ is more fit than $s_i$, the most fit seed in generation $i$, where $i < n$ and $p_{in} \in [0,1]$. We estimate $p_{in}$ by having $s_i$ and $s_n$ compete against each other in $g$ Immigration Games. In the experiments that follow, $g$ is set to 50 games. If $s_n$ wins $w$ games against $s_i$, then we estimate the probability $p_{in}$ by $w/g$. If $s_i$ and $s_n$ are equally fit, we expect $p_{in} = 0.5$. To satisfy the first requirement for an external, absolute fitness measure (directional consistency), the curve should rise when $p_{in} > 0.5$, fall when $p_{in} < 0.5$, and stay flat when $p_{in} = 0.5$. To achieve this behaviour, we use the formula $2p_{in} - 1$, which ranges from $-1$ to $+1$ as $p_{in}$ ranges from 0 to 1. The external, absolute fitness $f_n$ of $s_n$ is then defined as follows:

$$f_n = \sum_{i=0}^{i=n-1} (2p_{in} - 1)$$

The function $f_n$ ranges from $-n$ to $+n$. The function has directional consistency: If $p_{in}$ reaches a generation $n$ where the probability of winning is random ($p_{in} = 0.5$), then the curve for $f_n$ will start to flatten out. If the probability is worse than random ($p_{in} < 0.5$), the curve will head downwards, perhaps eventually going below zero. If the probability is better than random ($p_{in} > 0.5$), the curve will head upwards. The function also has slope consistency: The slope of the curve corresponds to the pace of fitness change. Thus, this function satisfies the two requirements for an external, absolute fitness measure.

We reduce noise in our estimate of the probability $p_{in}$ by taking the top ten fittest seeds in generation $i$ and the top ten fittest seeds in generation $n$ and making each pair of seeds compete twice, so that the estimate for $p_{in}$ is based on the average outcome of 200 competitions (10×10×2 = 200). The noise is further reduced by averaging over twelve separate runs of the model.

Figure 11 shows the fitness of the six layers, as given by the new fitness measure. The new fitness measure makes the steady fitness increase of Layer 4 more readily visible than the old fitness measure (compare Figure 11 with Figures 6 and 10).

Insert Figure 11 here.

Table 9 shows that the two external fitness measures, comparison with random seeds (Figures 6 and 10) and comparison with past winners (Figure 11) are highly correlated (0.767) when we consider the fitness score $f_n$ averaged over all generations. The correlation is statistically significant.

Insert Table 9 here.

Table 10 shows that the two external fitness measures are also highly correlated (0.765) when we focus on the final generation ($n = 100$) and the correlation is again statistically significant. All three fitness measures (in Sections 4.1, 4.3, and 4.4) show the same general rankings of the six different configurations of Model-S: (a) Layer 1 has the lowest fitness, (b) Layers 2 and 3 are similar, with a slight advantage to Layer 2, and (c) Layer 4, Layer 4 Shuffled, and Layer 4 Mutualism have the highest fitness and are similar.

Insert Table 10 here.

Open-ended evolution is defined as evolution that is [25, page 409] "capable of producing a continual stream of novel organisms rather than settling on some quasi-stable state beyond which nothing fundamentally new occurs." In Figure 11, it appears that evolution is open-ended over the course of 100 generations for Layer 4, Layer 4 Shuffled, and Layer 4 Mutualism. We hypothesize that the trends in Figure 11 will continue indefinitely.

## 5 Discussion of Results

Section 4.1 shows that Layer 4, symbiosis by genetic fusion, significantly increases fitness when combined with the other layers. The increase in fitness due to fusion occurs despite several obstacles: (1) Fusion is rare, taking place in only 1 out of 200 births (prob_fusion = 0.005; see Table 1). (2) Fission is twice as likely as fusion (prob_fission = 0.01) and fission is pushing the population towards decreasing size, in opposition to fusion. (3) Layer 4 passes control to Layer 3 most of the time (98.5% of the time; see Figure 5), but Layer 3 is less fit than Layer 2 (see Figure 6). Layer 4 could perform better if it passed control to directly to Layer 2 instead of Layer 3. (4) The parameters max_area_first and max_area_last impose strong constraints on fusion (see Section 3.5). When the population reaches the limits set by these parameters, fusion is no longer permitted (because the simulation becomes exceedingly slow).

Section 4.2 considers whether the fitness increase from fusion is due to increase in the size of the fused seeds or due to beneficial interaction of the fused seeds. Shuffling is introduced as a way to disrupt the structure of a seed without affecting its size. The shape, area, and density of a seed are not affected by shuffling. The results indicate that shuffling reduces the fitness of the fused seed (see Table 6), which slows the increase in fitness in the earlier generations, but the population recovers in the later generations (see Figure 10 and Table 5), due to the limits on size (max_area_first and max_area_last in Section 3.5).

Section 4.2 also examines what happens when fusion requires mutual benefit, with Layer 4 Mutualism. The results show that the few cases of mutual benefit account for all of the increasing fitness. Eliminating all of the fusion events that lack mutual benefit has no impact on fitness: There is no significant difference between the fitness curves for Layer 4 (symbiosis as any association) and Layer 4 Mutualism (symbiosis as persistent mutualism) in Figures 10 and 11.

Section 4.3 shows that comparison of evolved seeds with human-engineered seeds (Table 8) yields approximately the same ranking of the layers of Model-S as comparison with random seeds of the same size and density (Table 7). This agreement between fitness measured by comparison with random seeds (Section 4.1) and fitness measured by comparison with human-engineered seeds (Section 4.3) suggests that both fitness measures are performing as intended.

Section 4.4 introduces a third absolute, external measure of fitness, in addition to the measures in Sections 4.1 and 4.3. This fitness measure is designed to have both *directional consistency* (it rises when the fitness of the population is improving, stays flat when the population is neither improving nor worsening, and falls when the population is worsening) and *slope consistency* (the pace of fitness change corresponds to the slope of the curve). The measure is based on comparing the most fit seed in each generation with the most fit seed in all previous generations (where the most fit seed is determined by the internal, relative fitness of the seeds in the given population). Tables 9 and 10 show that this new measure of fitness is highly correlated with external fitness measured by comparison with random seeds. Furthermore, the steady increase in the fitness of Layer 4, Layer 4 Shuffled, and Layer 4 Mutualism that we see with this third fitness measure (Figure 11) lends support to the hypothesis that symbiosis supports open-ended evolution.

Three measures of external, absolute fitness all produce the same general ranking of the six layers: (a) comparison of fit seeds with random seeds of the same size and density in Figures 6 and 10, Sections 4.1 and 4.2, (b) comparison of fit seeds with human-designed seeds in Table 8, Section 4.3, and (c) comparison of current fit seeds with past fit seeds in Figure 11, Section 4.4. The agreement among these three different measures shows the results are not dependent on a specific approach to measuring external, absolute fitness.

Three different arguments support the claim that increasing area is *necessary* but *not sufficient* for increasing fitness, thus *increasing area does not cause increasing fitness*: (1) The argument at the end of Section 4.1, comparing evolved seeds with area-matched random seeds, shows that Layer 4 generates seeds with highly non-random structures, and the difference in fitness between the evolved seeds and the random seeds cannot be due to a difference in area, because the competing seeds have the same area. (2) The argument in Section 4.2, comparing Layer 4 and Layer 4 Mutualism, shows there is no significant difference in their fitness curves (Figure 10), although Layer 4 produces many more large seeds than Layer 4 Mutualism (Table 4). The relative abundance of large seeds does not boost the fitness of Layer 4, because most of the large seeds lack mutualism. (3) The argument in Section 4.2, comparing Layer 4 and Layer 4 Shuffled, shows that shuffling reduces fitness (Tables 5 and 6), although it does not change area. The cause of the reduction in fitness must be structural, since the only difference in the seeds is structural.

The necessary and sufficient condition for increasing fitness is the evolution of structures that are increasingly better at playing the Immigration Game. For unbounded fitness increase, increasing area is necessary to accommodate increasingly complex structures, but increasing area is not sufficient.

# 6 Future Work and Limitations

In biology, there are many examples of symbiosis without mutualism, such as parasitism. Model-S does not currently deal with parasitism, although it could be expanded to model parasitism. We should not infer from the experiments described here that evolution favours mutualism over parasitism. The genetic fusion operator in Model-S encourages the two fused genomes to get along with each other, because selection occurs at the level of the whole. If a parasite kills its host, selection at the level of the whole would penalize the parasite. Genetic fusion is not the appropriate operator for modeling the parasite-host relationship. A parasite and its host have a close relationship, but their genes are selected separately. Fusion is essentially a mechanism for forcing the parts to either cooperate or die. We leave parasitism for future work.

In Section 4.2, we examined the interaction between the two parts of a fused seed. Fusion is a joining of two seeds (two genomes), which are inert, so there is no interaction at the moment of fusion. However, there is vigorous interaction when the Immigration Game runs (the growth of the two phenomes). Symbiosis occurs during the game, either *symbiosis as any association* or *symbiosis as persistent mutualism*, depending on how well the seeds work together. We can easily detect the interaction by comparing the fitness of the separate parts with the fitness of the fused whole. Table 6 tells us that the interaction is mostly negative; only 15% of the time do both parts of the fused whole benefit (mutualism). Mostly we see association without benefit. This is expected for the fusion operator, just as it is for the mutation operator: Most mutations are harmful or neutral, only a few are beneficial, but it is those few beneficial cases that yield greater fitness and adaptation in the long-term evolution of the population.

We can infer the interaction between the two parts of a fused seed from the impact on fitness, but it would be interesting to observe the interaction directly. In future work, we could visually display the interaction between the two fused seeds. For example, suppose that a red seed that has undergone fusion is competing with a blue seed that has undergone fusion. We could colour the red seed in two different shades of red, corresponding to its two fused parts, and we could colour the blue seed in two different shades of blue, corresponding to its two fused parts. We would then be able to visualize how the parts interact over the course of a game. (We need to decide how to handle the case where an empty cell has three live neighbours with three different colours. There are many ways to deal with this.)

A limitation of Model-S is the amount of time required to run the Immigration Game as the seeds evolve to become larger with fusion. Addressing this problem may be a straightforward task of tuning the parameters, width_factor, height_factor, and time_factor (see Section 3.2), but we believe a more sophisticated method is required for determining the best toroid size and the best time limit for a given pair of competing seeds. One way to set the time limit would be to use a test for *quiescence* to determine the end of a game. The idea is to end the game when the score appears to be nearly stable.

A puzzle from the results presented in Section 4 is the relatively poor performance of sexual reproduction in Layer 3. Simon [21] lists eleven different kinds of genetic crossover. It may be that one of the other forms of crossover will perform better than the simple single-point crossover used in Model-S. It might be helpful to introduce a form of two-dimensional crossover that exchanges sub-squares between the parent seed matrices.

We use a form of restricted mating in Model-S, controlled by the min_similarity and max_similarity parameters (see Section 3.4). We tried to tune these parameters to improve sexual reproduction, without success. Sexual reproduction has long been a topic for debate among evolutionary biologists, with many different theories about its role in evolution. Ridley [20] argues that a major reason for sexual reproduction is to provide resistance against parasites. One option would be to add simulated parasites to Model-S.

Table 1 shows that the parameter space for Model-S is relatively large. Experiments with Model-S are relatively slow, which makes it difficult to explore the parameter space thoroughly. Although we have run many experiments, we have only explored a tiny fraction of the parameter space. Much exploration remains to be done.

For those who are interested in Lamarckian evolution, Model-S could be a suitable platform. Lamarckian evolution is based on the inheritance of acquired characteristics. We can simulate a kind of Lamarckian evolution as follows: (1) Put a seed into the Game of Life. This seed is the genotype. (2) Let the game run for *N* steps. The resulting pattern is the phenotype. (3) Take the resulting pattern out of the game and use it as a new seed. This new seed has acquired characteristics from its time in the Game of Life and these characteristics are heritable.

For those who are interested in tracking the heritage of individuals, it would be easy to modify Model-S by storing a family tree in each seed object. The nodes in the tree could be pointers into a database of stored seeds. This would be useful for testing hypotheses about the properties of inheritance in Model-S.

The discussion of related work (Section 2) mentioned the meta-model of Banzhaf et al. [2], which can be applied to the Game of Life. An interesting project for future work would be to apply the meta-model to the Immigration Game and Model-S.

# 7 Conclusion

Our model of symbiosis has four layers of genetic operators. The first three layers include asexual and sexual reproduction, with standard genetic operators such as mutation and crossover. The fourth layer introduces two new genetic operators, fusion and fission.

In the model, the fitness of an organism is determined by competition in the Immigration Game, a variation on the Game of Life. A key insight is that the fusion operator is easy to implement in the Game of Life: Organisms are fused by simply joining them side-by-side and treating them as a new whole. In other types of artificial life simulations, fusion may not be as straightforward to implement.

Our main result is that symbiosis by fusion is a powerful genetic operator, when combined with the standard genetic operators (mutation and crossover). A small amount of fusion (one birth in 200) can have a substantial impact on the course of evolution (Section 4.1). The results with the new unbounded external fitness measure (Section 4.4, Figure 11) suggest that fusion can sustain open-ended evolution [25].

We hope that the release of Model-S as open-source software [30] will encourage other researchers to explore the many open questions raised in this paper. It seems likely that there are many other genetic operators, beyond mutation, selection, crossover, fission, and fusion, awaiting artificial life models.

## Acknowledgments

Thanks to Tim Taylor and Martin Brooks for helpful discussion and advice. Thanks to Andrew Trevorrow, Tom Rokicki, Tim Hutton, Dave Greene, Jason Summers, Maks Verver, Robert Munafo, Brenton Bostick, and Chris Rowett, for developing Golly, which made this research more productive and enjoyable. Thanks to the *Artificial Life* reviewers for their very thoughtful and helpful comments.

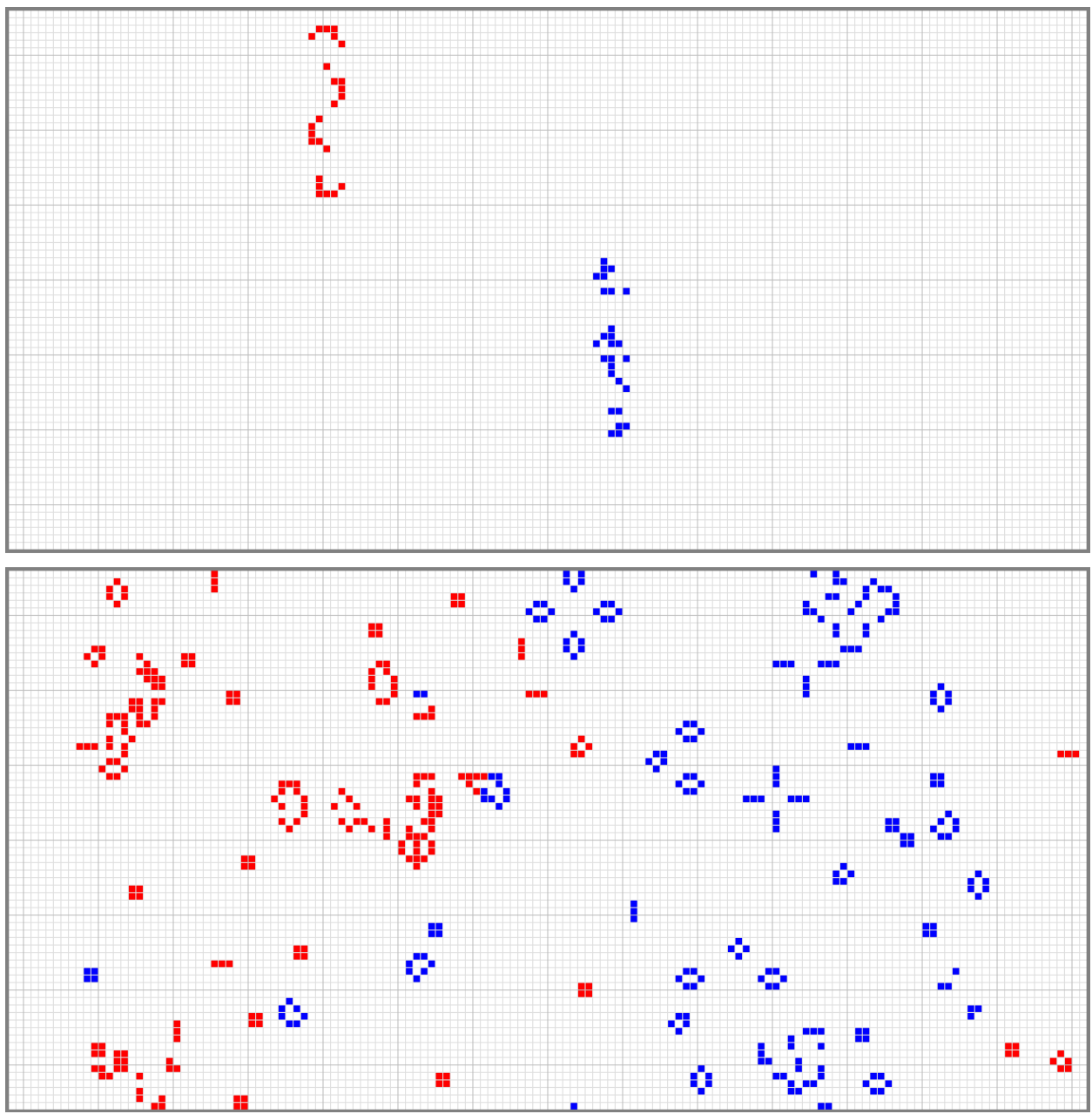

Figure 1. The first image above shows the initial state of an Immigration Game and the second image shows the final state, when the game reached its time limit. The first image contains two competing seeds, a red seed (24 × 5 block, 24 live cells, density 0.200) and a blue seed (24 × 5 block, 27 live cells, density 0.225). The second image reveals that blue won the game. Blue grew by 209 live cells (from 27 to 236) and red grew by 204 live cells (from 24 to 228). Both seeds were the fittest seeds in the final generations of two different runs of Model-S, in which both runs used all four layers of the model. The second image is a typical example of the final state of a game.

[See Section 1]

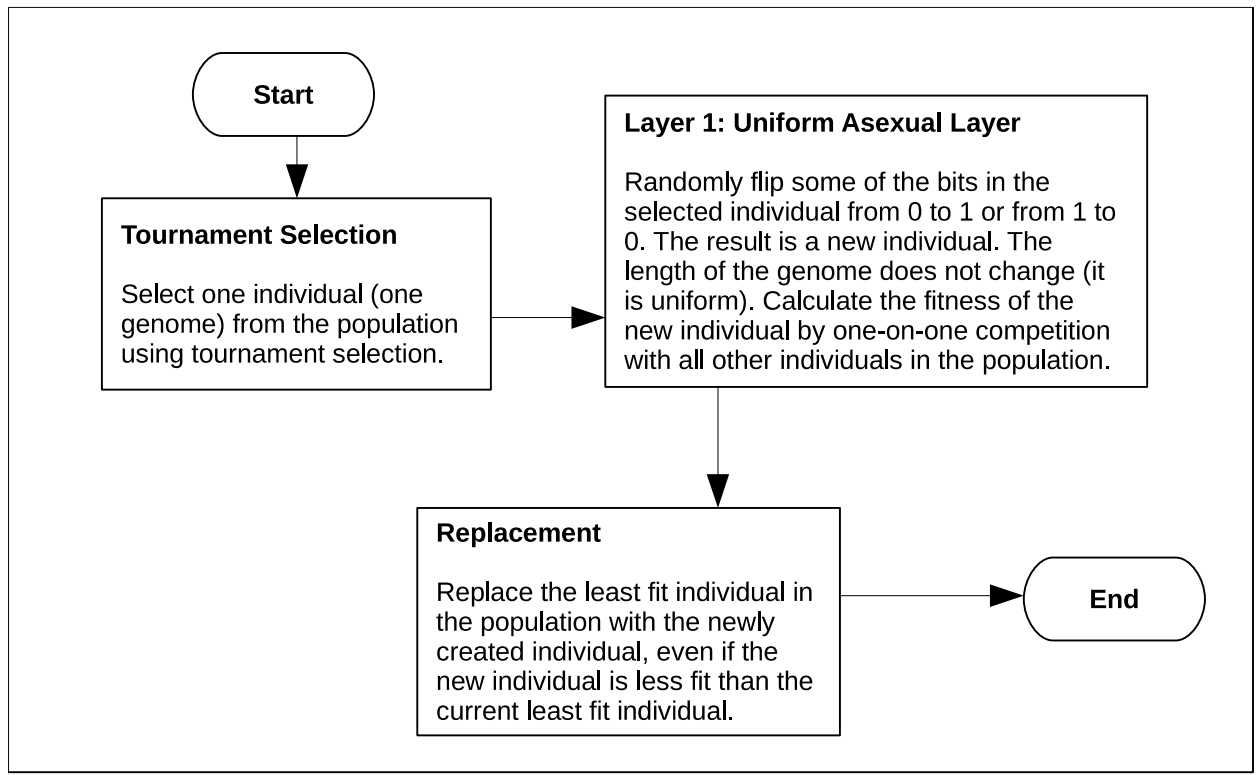

Figure 2. The flowchart above describes the process for uniform asexual reproduction. This process is a subroutine in a loop that produces a series of new individuals. For each individual that is added to the population, another is removed; hence this is a steady-state model with a constant population size. Uniform asexual reproduction takes the input individual and generates a mutated copy as the output.

[See Section 3.2]

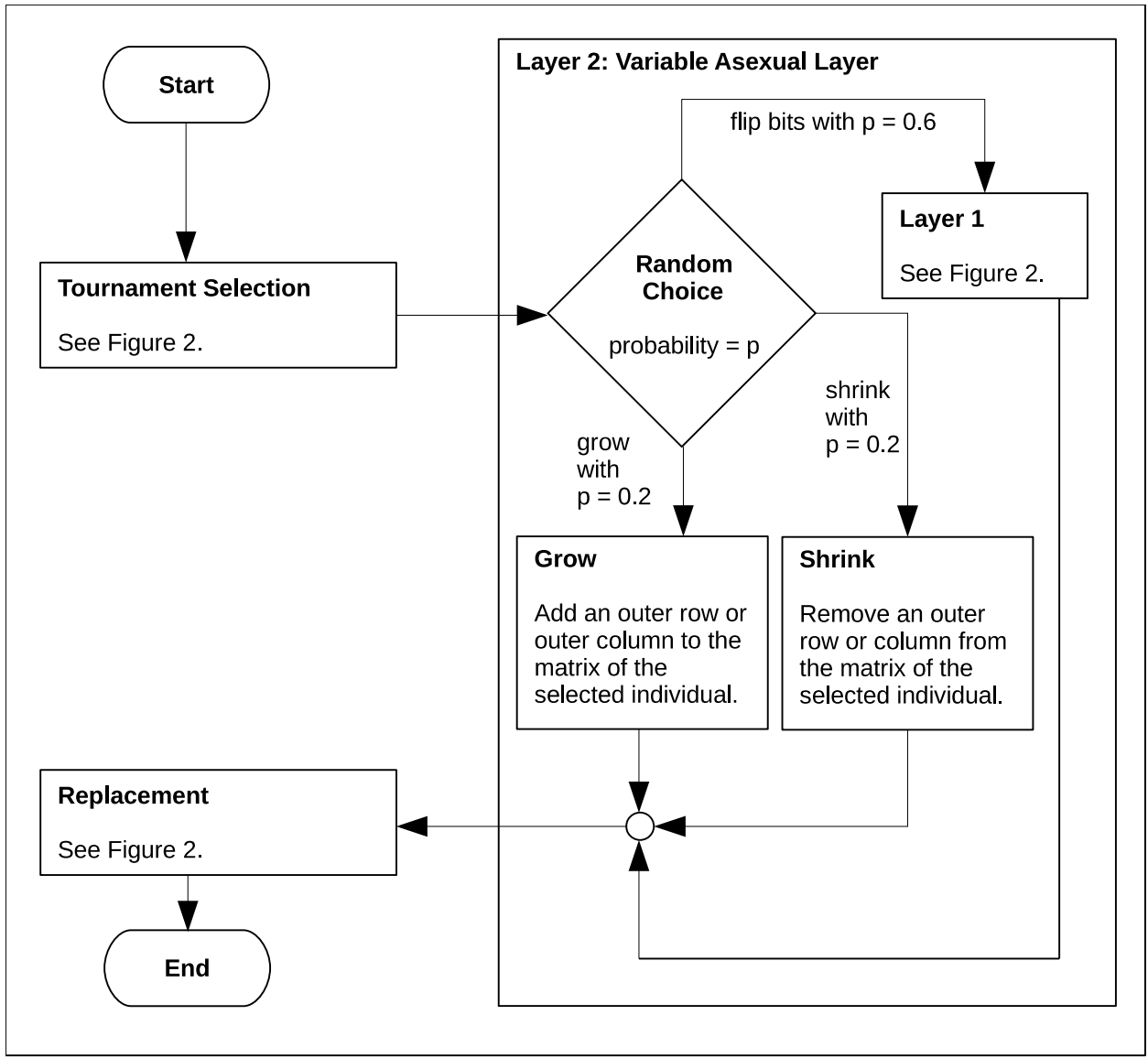

Figure 3. Layer 1 produces individuals of uniform size, whereas Layer 2 allows variable size. A random number is generated between 0 and 1. The value of the random number determines whether the individual will shrink in size, grow in size, or be passed on to Layer 1, where it will copy the size of its parent and mutate by flipping bit values.

[See Section 3.3]

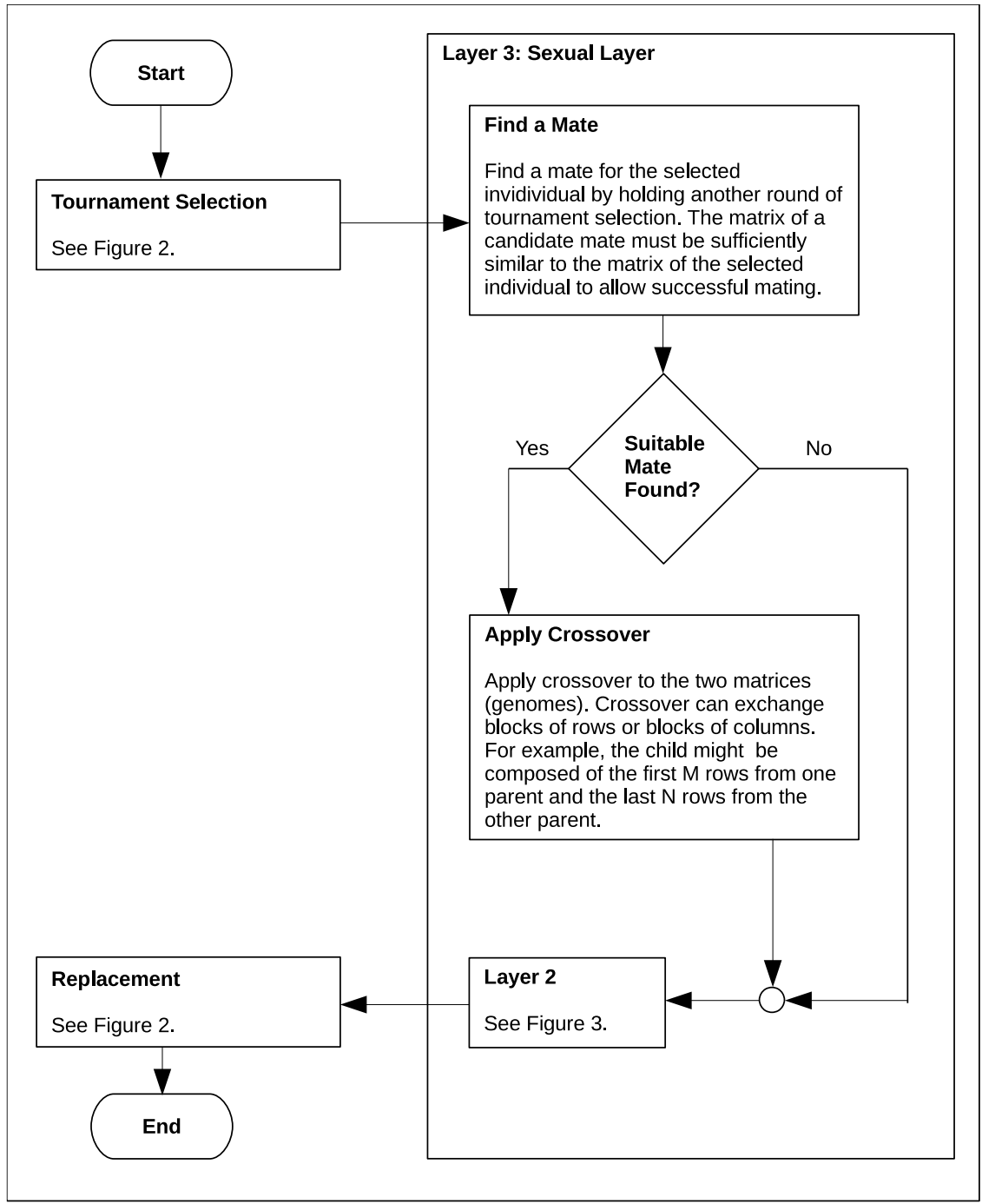

Figure 4: Layer 3 introduces sexual reproduction, where part of one individual's matrix is combined with part of another individual's matrix. The individuals are neither male nor female; any individual can mate with any other individual, so long as they are sufficiently similar. After mating, the child individual is passed on to Layer 2 where it grows, shrinks, or flips bits.

[See Section 3.4]

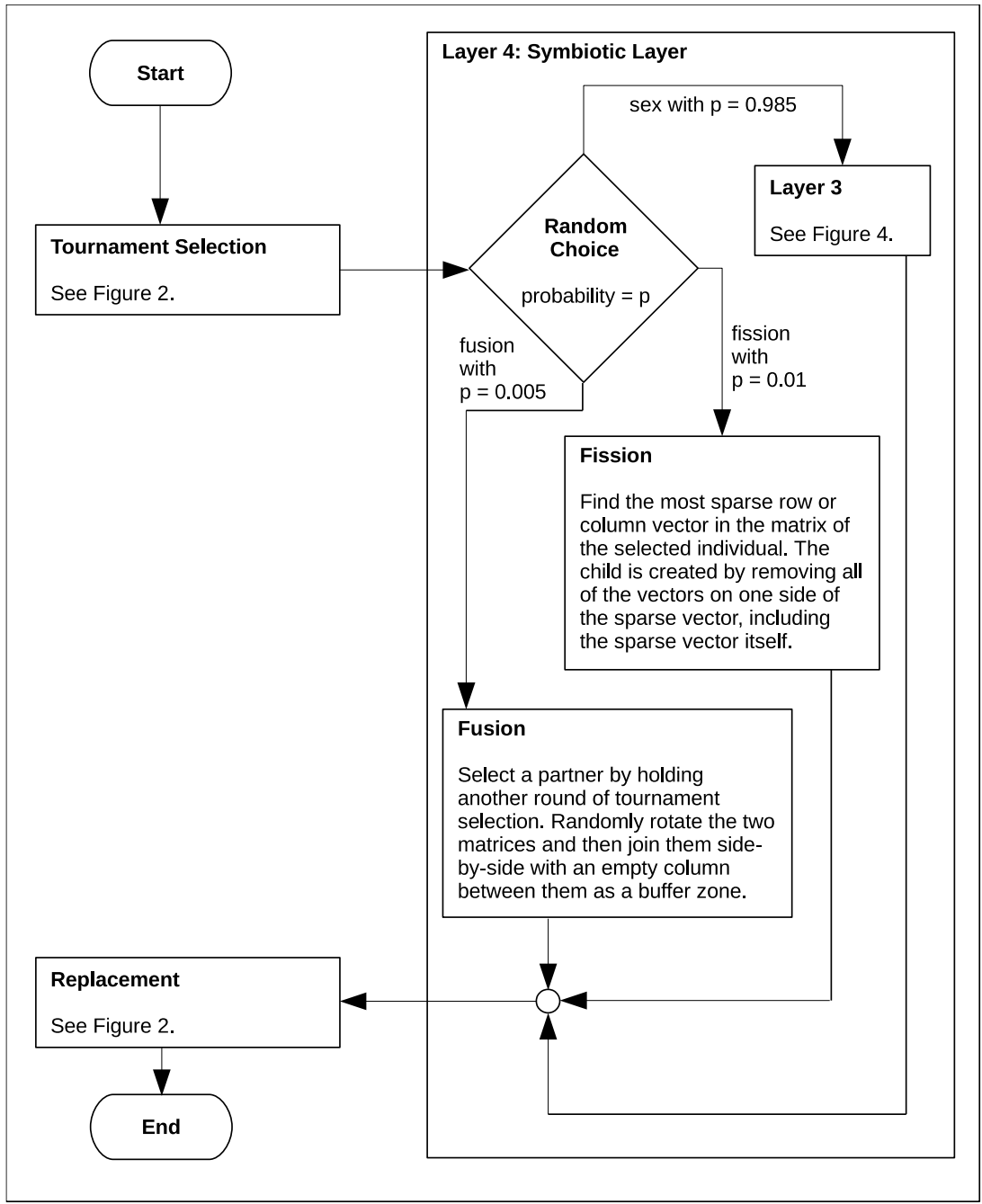

Figure 5. Unlike Layers 1, 2, and 3, Layer 4 is not a form of reproduction. With fusion, a new individual is created by fusing two individuals. With fission, a new individual is created by breaking an individual into two parts. Only one of the two parts is kept. We set the probability of fission higher than the probability of fusion in order to see whether selection can overcome this bias towards fission. Note that fission and fusion are much less likely than sexual reproduction.

[See Section 3.5]

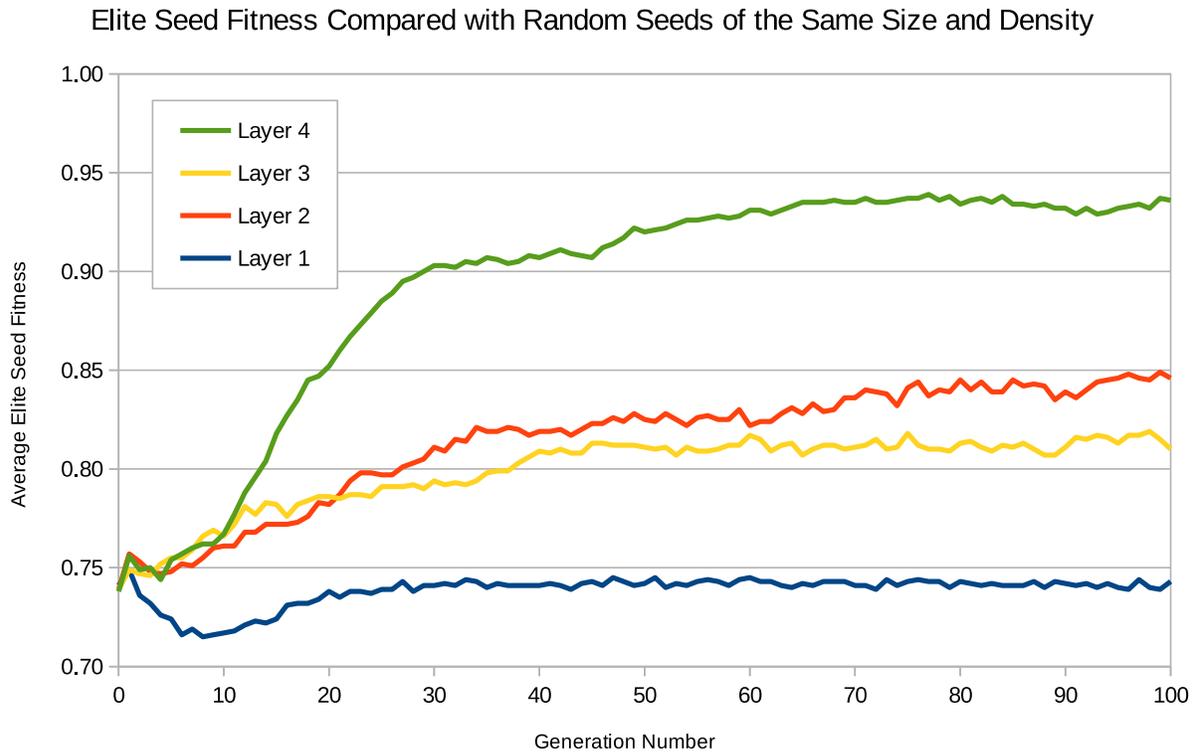

Figure 6. Each curve in this figure (that is, each layer) is the average of 12 separate runs of Model-S. The fitness of a seed is the fraction of Immigration Game contests that it wins when competing against randomly generated seeds with the same size (the same width and height) and the same density (the same number of live cells). This is an external measure of fitness that does not correspond to the internal measure used in the selection process in the four layers. The internal measure of fitness would show no progress, because it compares each individual to the population, and the population as a whole is progressing (on average) as fast as the individuals in the population are progressing.

[See Section 4.1]

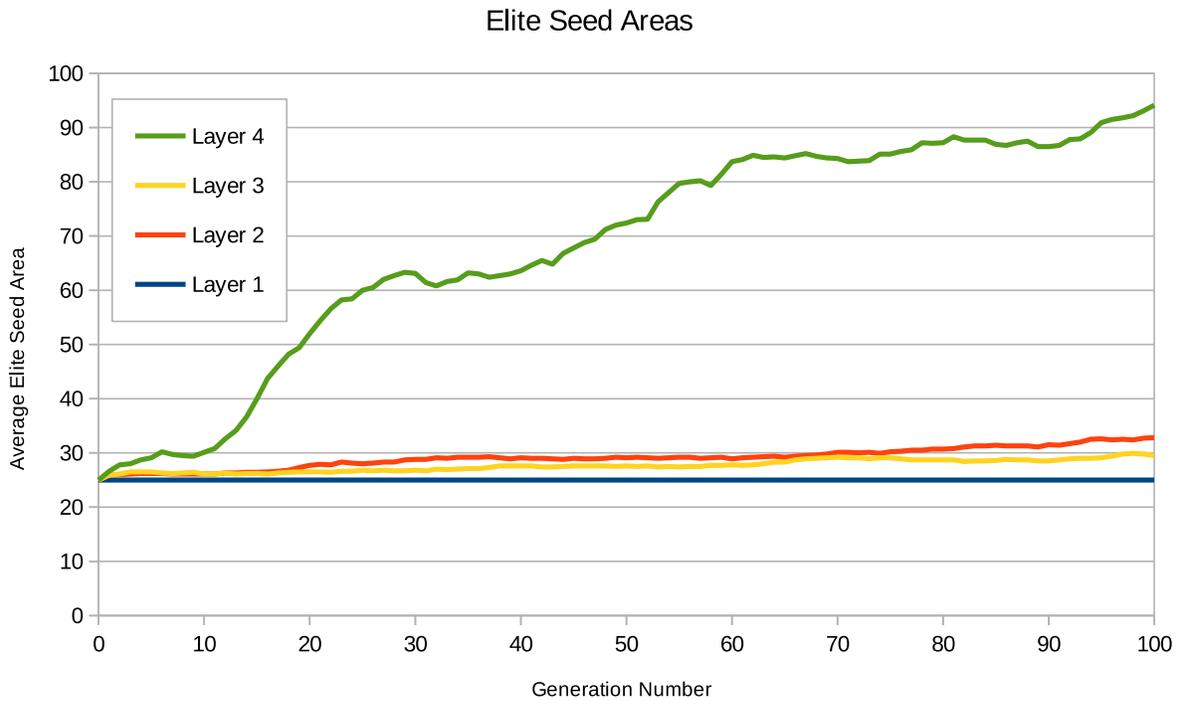

Figure 7. Each curve in this figure is the average of 12 separate runs of Model-S. Comparing this figure with Figure 6 suggests that area and fitness are positively correlated, even though the fitness in Figure 6 is based on size-matched competitions. Greater area indirectly helps fitness by allowing more information to be encoded, which permits more complex structures and actions.

[See Section 4.1]

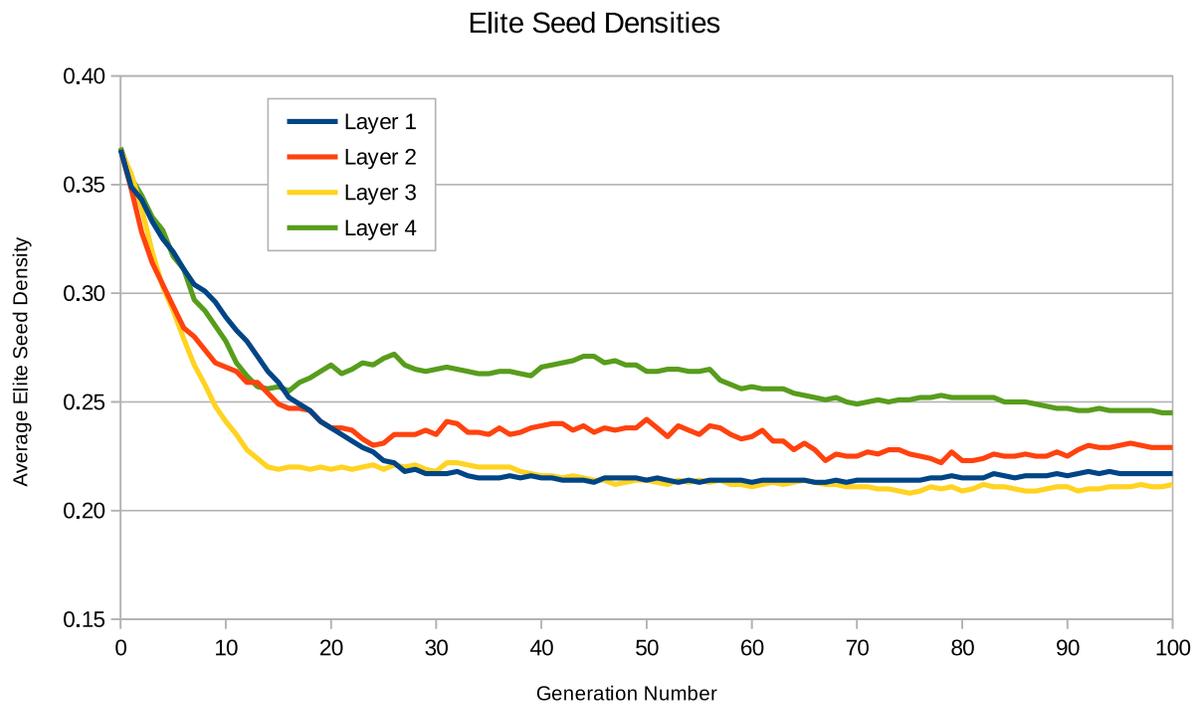

Figure 8. In all four layers, density decreases over time. Perhaps lower density allows information to travel longer distances, enabling greater complexity.

[See Section 4.1]

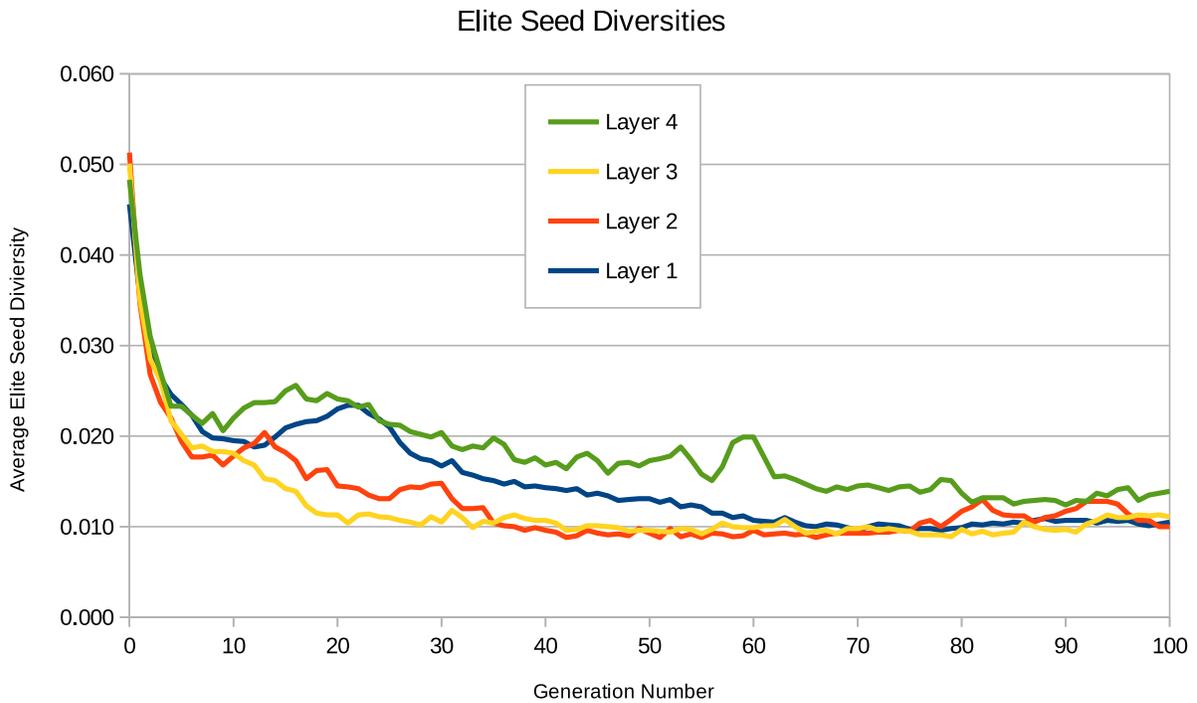

Figure 9. The diversity of the elite seed population is measured here by the standard deviation of the external measure of fitness (see Figure 6). The standard deviation is then averaged over the 12 separate runs of Model-S for each layer. Layers 1, 2, and 3 appear to have approximately the same diversities, but Layer 4 appears to be more diverse.

[See Section 4.1]

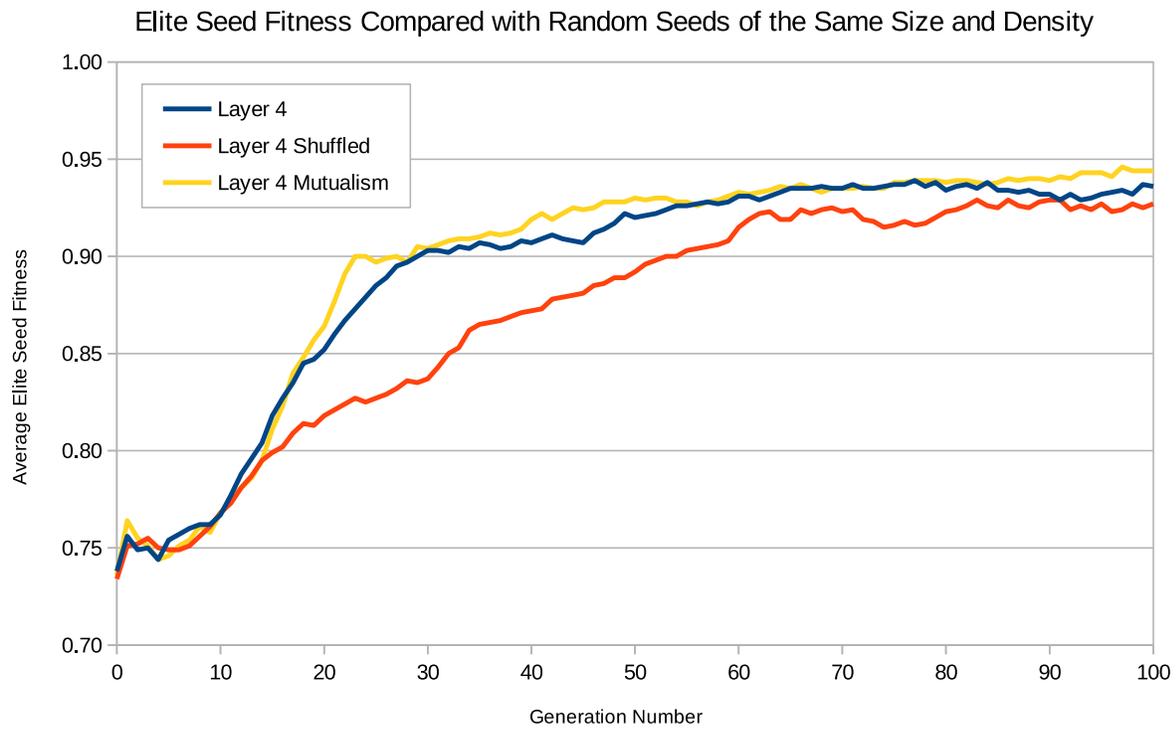

Figure 10. (1) Would fusion work equally well if it combined one evolved seed and one random seed? Shuffling one of the two seeds before fusing them (Layer 4 Shuffled) reduces the fitness of the fused seeds, compared to fusing seeds without shuffling (Layer 4 and Layer 4 Mutualism). (2) What happens when we model symbiosis as persistent mutualism? There is no significant difference between *symbiosis as any association* (Layer 4) and *symbiosis as persistent mutualism* (Layer 4 Mutualism). This indicates that the fitness increase over the generations comes from the cases where symbiosis is mutually beneficial.

[See Section 4.2]

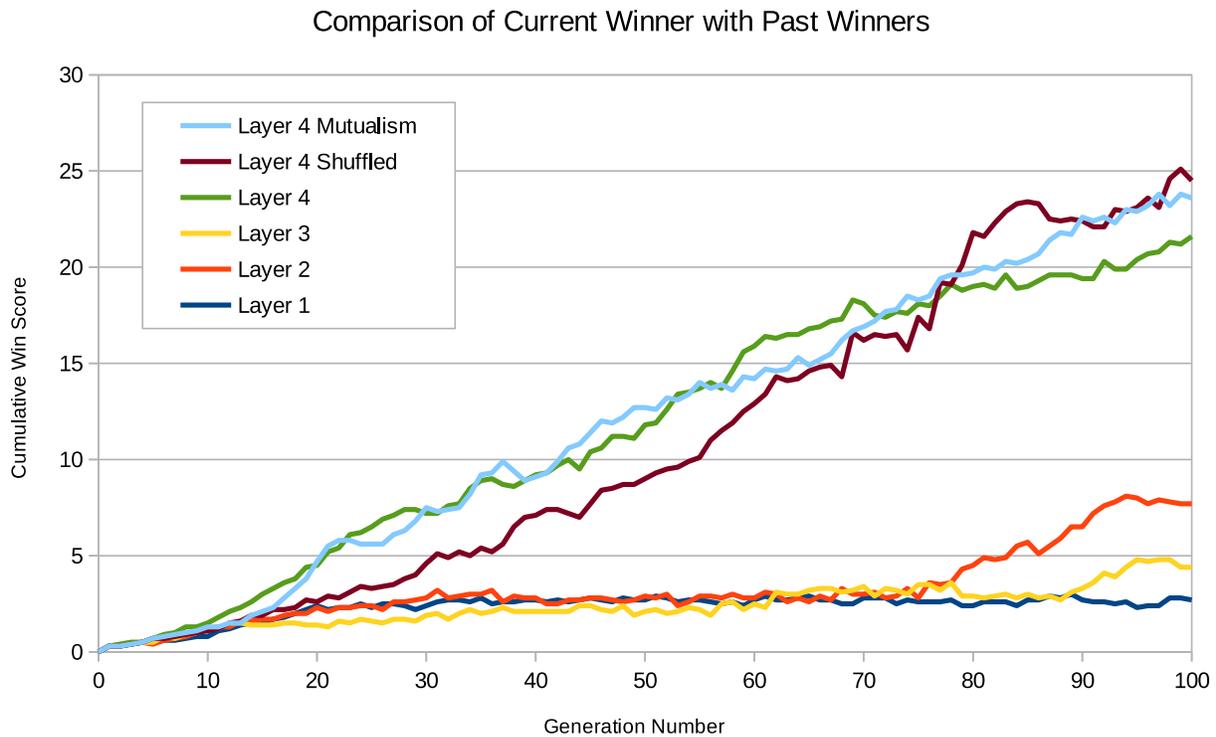

Figure 11. This figure compares the six different configurations of Model-S using an unbounded external fitness measure. Whereas the fitness measure in Figures 6 and 10 (comparing evolved seeds with random seeds of the same size and density) is limited to ranging from 0 to 1, the fitness measure here ranges from negative infinity to positive infinity. Comparing Figures 6 and 11, we see the same ranking of the different layers (from top to bottom: Layer 4, Layer 2, Layer 3, Layer 1). Figure 11 is more suitable than Figure 6 for showing the steady increase in fitness of Layer 4. In Figure 11, as in Figure 10, Layer 4 Shuffled falls behind Layer 4 and Layer 4 Mutualism, but it eventually catches up.

[See Section 4.4]

Table 1. This table lists the parameters used in Model-S and their values for the experiments presented in Section 4.1. One generation is defined as the birth of pop_size children, therefore the number of children born in one run is pop_size × num_generations = 20,000. The number of runs is not a parameter in the model; each run begins by starting a new instance of Golly.

| Parameter Names | Parameter Values | Used in Layers |
|---|---|---|
| experiment_type_num | 1, 2, 3, 4 | 1, 2, 3, 4 |
| pop_size | 200 | 1, 2, 3, 4 |
| num_trials | 2 | 1, 2, 3, 4 |
| num_generations | 100 | 1, 2, 3, 4 |
| min_s_xspan | 5 | 1, 2, 3, 4 |
| min_s_yspan | 5 | 1, 2, 3, 4 |
| s_xspan | 5 | 1, 2, 3, 4 |
| s_yspan | 5 | 1, 2, 3, 4 |
| max_area_first | 120 | 1, 2, 3, 4 |
| max_area_last | 170 | 1, 2, 3, 4 |
| seed_density | 0.375 | 1, 2, 3, 4 |
| width_factor | 6.0 | 1, 2, 3, 4 |
| height_factor | 3.0 | 1, 2, 3, 4 |
| time_factor | 6.0 | 1, 2, 3, 4 |
| tournament_size | 2 | 1, 2, 3, 4 |
| elite_size | 50 | 1, 2, 3, 4 |
| mutation_rate | 0.01 | 1, 2, 3, 4 |
| prob_flip | 0.6 | 2, 3, 4 |
| prob_grow | 0.2 | 2, 3, 4 |
| prob_shrink | 0.2 | 2, 3, 4 |
| min_similarity | 0.8 | 3, 4 |
| max_similarity | 0.99 | 3, 4 |
| prob_fission | 0.01 | 4 |
| prob_fusion | 0.005 | 4 |
| symbiosis_flag | 0 | 4 |
| fusion_test_flag | 0 | 4 |

[See Section 4.1]

Table 2. This table tests the statistical significance of the differences in the fitness curves in Figure 6. Each layer (each curve) is summarized by the average fitness over a run, yielding a sample of twelve values, one value for each of the twelve runs of a layer. We then compare the fitness curves for all possible pairs of layers, using a two-tailed Welch t-test for samples with unequal variance (heteroscedastic variance). All the pairs of curves in Figure 6 are significantly different, except for layers 2 and 3 (the variable asexual layer and the sexual layer).

| Layers to Compare | p-value | p-value $< 0.05$ |
|---|---|---|
| Layer 1 vs Layer 2 | 5.03E-06 | Yes |
| Layer 1 vs Layer 3 | 4.10E-05 | Yes |
| Layer 1 vs Layer 4 | 2.81E-10 | Yes |
| Layer 2 vs Layer 3 | 2.79E-01 | No |
| Layer 2 vs Layer 4 | 3.44E-06 | Yes |
| Layer 3 vs Layer 4 | 2.33E-07 | Yes |

[See Section 4.1]

Table 3. This table looks at the correlations between fitness, area, density, and diversity (as displayed in Figures 6 to 9). We evaluate the statistical significance of the correlations using a two-tailed Student t-test for Pearson correlations. All the correlations are statistically significant. Each correlation is based on comparing two samples of 48 values each (whereas Table 2 compares two samples of 12 values each).

| Feature 1 | Feature 2 | Correlation | p-value  | p-value < 0.05 |
|-----------|-----------|-------------|----------|----------------|
| area      | density   | 0.362       | 1.14E-02 | Yes            |
| area      | diversity | 0.816       | 1.64E-12 | Yes            |
| area      | fitness   | 0.843       | 5.89E-14 | Yes            |
| density   | diversity | 0.418       | 3.11E-03 | Yes            |
| density   | fitness   | 0.405       | 4.28E-03 | Yes            |
| diversity | fitness   | 0.566       | 2.72E-05 | Yes            |

[See Section 4.1]

Table 4: This table compares fusion events during runs of Layer 4, Layer 4 Shuffled, and Layer 4 Mutualism. The numbers are averages over the 12 runs of each layer. The column *Number of Fusion Events* includes events with and without mutualism. The other columns focus exclusively on fusion events with mutualism. By design, all fusion events in Layer 4 Mutualism are cases of mutualism. The expected number of fusion events is prob_fusion × num_generations × pop_size = 100, but fusion is suppressed by the limits max_area_first and max_area_last in Layer 4 and Layer 4 Shuffled, and fusion is further suppressed by the requirement of mutual benefit in Layer 4 Mutualism.

|                   | Number of Fusion Events | Number of Mutualisms | Percent of Mutualisms |
|-------------------|-------------------------|----------------------|-----------------------|
| Layer 4           | 44.3                    | 6.6                  | 15%                   |
| Layer 4 Shuffled  | 61.5                    | 4.9                  | 8%                    |
| Layer 4 Mutualism | 5.4                     | 5.4                  | 100%                  |

[See Section 4.2]

Table 5: This table compares Layer 4 with Layer 4 Shuffled in generations 30 and 100. The fitness of the two layers for all generations is shown in Figure 10. Layer 4 is significantly more fit than Layer 4 Shuffled in generation 30, but the difference between the two layers is no longer significant when they reach generation 100. Averaged over all the generations, the difference is not significant. The p-values are calculated using a two-tailed Welch t-test for samples with unequal variance (heteroscedastic variance).

| Generation | Layer 4 | Layer 4 Shuffled | p-value | p-value $< 0.05$ |
|---|---|---|---|---|
| 30 | 0.903 | 0.837 | 0.004 | Yes |
| 100 | 0.936 | 0.927 | 0.601 | No |
| All | 0.892 | 0.871 | 0.057 | No |

[See Section 4.2]

Table 6: In this table, we compare Layer 4 with Layer 4 Shuffled. Shuffling decreases the probability that the component parts will both benefit from fusion; that is, the probability of *symbiosis as persistent mutualism*. The whole is more fit than both parts 15% of the time for Layer 4, but only 8% of the time for Layer 4 Shuffled. This shows that increased fitness does not come merely from the increase in area when fusing two seeds; a significant part of the increased fitness comes from the structural properties of the two seeds that are fused together. Layer 4 Shuffled is less likely to have the proper structure than Layer 4. The p-values in the table are calculated using a two-tailed Welch t-test for samples with unequal variance (heteroscedastic variance).

|  | No Parts Benefit | One Part Benefits | Both Parts Benefit |
| --- | --- | --- | --- |
| Layer 4 | 80.5% | 4.5% | 15.0% |
| Layer 4 Shuffled | 89.3% | 2.7% | 8.0% |
| p-value of Difference | 0.041 | 0.214 | 0.033 |
| p-value $< 0.05$ | Yes | No | Yes |

[See Section 4.2]

Table 7. This table summarizes the final generations of the layers (generation 100). The average area of the evolved seeds (60.5 in this table) is generally smaller than the area of the human-designed seeds (2454 in Table 8). The evolved seeds are competing against human-designed seeds that are mostly much larger than them.

| Layer | Fitness | Area | Density | Diversity |
|---|---|---|---|---|
| Layer 1 | 0.743 | 25.0 | 0.217 | 0.0105 |
| Layer 2 | 0.846 | 32.8 | 0.229 | 0.0100 |
| Layer 3 | 0.810 | 29.5 | 0.212 | 0.0111 |
| Layer 4 | 0.936 | 94.1 | 0.245 | 0.0139 |
| Layer 4 Shuffled | 0.927 | 90.5 | 0.245 | 0.0140 |
| Layer 4 Mutualism | 0.944 | 91.3 | 0.233 | 0.0142 |
| Average | 0.868 | 60.5 | 0.230 | 0.0123 |

[See Section 4.3]

Table 8. This table gives the scores of each of the layers when competing against human-designed patterns of comparable area. We test the layers against all human-designed patterns that have an area of 10,000 or less, a total of 29 patterns. Each human-designed Golly pattern competes 20 times against the fittest seed in the final generation of each run of the given layer. For example, there are 12 runs for Layer 1, thus 240 ($12 \times 20$) Immigration Games are played with each human-designed Golly pattern. In the table, we report the percentage of competitions in which the evolved seed was the winner.

|  |  | Percentage of Games Won for each Layer | | | | | |
| --- | --- | --- | --- | --- | --- | --- | --- |
| Golly Pattern File Name | Area | Layer 1 | Layer 2 | Layer 3 | Layer 4 | Layer 4 Shuffled | Layer 4 Mutualism |
| agar-p3.rle | 3456 | 100 | 100 | 100 | 100 | 100 | 100 |
| herringbone-agar-p14.rle | 2304 | 95 | 95 | 93 | 96 | 97 | 97 |
| pulsars-in-tube.rle | 136 | 0 | 4 | 8 | 81 | 82 | 73 |
| spacefiller.rle | 1274 | 0 | 0 | 0 | 3 | 3 | 2 |
| vacuum-cleaner.rle | 8730 | 0 | 13 | 13 | 75 | 79 | 56 |
| acorn.lif | 21 | 58 | 57 | 59 | 81 | 72 | 79 |
| ark1.rle | 928 | 1 | 6 | 8 | 42 | 41 | 41 |
| ark2.rle | 2332 | 0 | 2 | 2 | 21 | 26 | 22 |
| blom.rle | 60 | 61 | 68 | 63 | 82 | 81 | 84 |
| iwona.rle | 420 | 11 | 27 | 28 | 67 | 69 | 64 |
| justyna.rle | 374 | 17 | 28 | 32 | 87 | 86 | 85 |
| lidka-predecessor.rle | 135 | 32 | 26 | 38 | 58 | 56 | 56 |
| natural-LWSS.rle | 40 | 65 | 73 | 69 | 79 | 78 | 80 |
| rabbits-relation-17423.rle | 36 | 66 | 72 | 73 | 86 | 85 | 84 |
| rabbits-relation-17465.rle | 24 | 75 | 80 | 76 | 81 | 85 | 86 |
| rabbits.lif | 21 | 69 | 71 | 71 | 85 | 82 | 85 |
| temp-pulsars-big-s.rle | 64 | 77 | 83 | 79 | 84 | 82 | 83 |
| die658.rle | 400 | 97 | 95 | 95 | 98 | 98 | 98 |
| line-puffer-superstable.rle | 4992 | 0 | 0 | 0 | 3 | 2 | 3 |
| line-puffer-unstable.rle | 1683 | 2 | 3 | 2 | 12 | 19 | 10 |
| pi-fuse-puffer.rle | 1827 | 18 | 36 | 42 | 90 | 89 | 83 |
| puffer-2c5.rle | 8400 | 1 | 5 | 8 | 63 | 57 | 44 |
| puffer-train.rle | 90 | 34 | 43 | 46 | 75 | 70 | 68 |
| heisenblinker-30.rle | 5032 | 17 | 27 | 31 | 80 | 75 | 72 |
| heisenburp-46-natural.rle | 2346 | 71 | 77 | 77 | 90 | 92 | 88 |
| eaters-misc.rle | 4851 | 100 | 98 | 99 | 96 | 96 | 98 |
| random.rle | 9604 | 100 | 100 | 100 | 100 | 100 | 100 |
| ss-eaters.rle | 7298 | 100 | 100 | 99 | 97 | 97 | 98 |
| stripey.rle | 4290 | 100 | 100 | 100 | 100 | 100 | 100 |
| Average | 2454 | 47 | 51 | 52 | 73 | 72 | 70 |

[See Section 4.3]

Table 9. This table compares the two forms of external fitness, elite seed fitness compared with random seeds of the same size and density (see Figures 6 and 10 in Sections 4.1 and 4.2) versus comparison of the current winner with past winners (see Figure 11 in Section 4.4). The table shows the correlation between the two measures of external fitness. The correlation is calculated from 72 random seed fitness scores and 72 past winner fitness scores (6 layers with 12 fitness values for each layer yields 72 fitness scores). Each fitness score in this table is an average over all the generations for a given layer (an average of 12 fitness values, one for each run, each of which is an average over the 100 generations). We evaluate the statistical significance of the correlations using a two-tailed Student t-test for Pearson correlations. The correlation is high (0.767) and it is statistically significant.

|  | Fitness Scores Averaged over all Generations | |
| --- | --- | --- |
|  | External Fitness Based on Comparison with Past Winners | External Fitness Based on Comparison with Random Seeds |
| Layer 1 | 2.3 | 0.739 |
| Layer 2 | 3.2 | 0.814 |
| Layer 3 | 2.4 | 0.799 |
| Layer 4 | 11.7 | 0.892 |
| Layer 4 Shuffled | 10.8 | 0.871 |
| Layer 4 Mutualism | 12.0 | 0.897 |
| Correlation | 0.767 | |
| p-value of Correlation | 4.15E-15 | |
| p-value $< 0.05$ | Yes | |

[See Section 4.4]

Table 10. Like the Table 9, this table compares the two forms of external fitness, elite seed fitness compared with random seeds of the same size and density versus comparison of the current winner with past winners. Each fitness score in this table is an average of the final generation for a given layer (an average of 12 fitness values). The difference between the two tables is that here we focus on the final generation, whereas the preceding table examined the average fitness over all the generations. The similar correlation values in Table 8 (0.767) and Table 9 (0.765) indicate that the correlations between the two forms of external fitness (comparison with past winners and comparison with random seeds) are robust.

|  | Fitness Scores in Generation 100 ||
|---|---|---|
|  | External Fitness Based on Comparison with Past Winners | External Fitness Based on Comparison with Random Seeds |
| Layer 1 | 2.7 | 0.743 |
| Layer 2 | 7.7 | 0.846 |
| Layer 3 | 4.4 | 0.810 |
| Layer 4 | 21.6 | 0.936 |
| Layer 4 Shuffled | 24.5 | 0.927 |
| Layer 4 Mutualism | 23.6 | 0.944 |
| Correlation | 0.765 ||
| p-value of Correlation | 5.37E-15 ||
| p-value < 0.05 | Yes ||

[See Section 4.4]